\documentclass[lettersize,journal]{IEEEtran}
\usepackage{amsmath,amsfonts}
\usepackage{amsmath}
\usepackage{algorithm}
\usepackage{algorithmic}
\usepackage{array}
\usepackage[caption=false,font=normalsize,labelfont=sf,textfont=sf]{subfig}
\usepackage{textcomp}
\usepackage{stfloats}
\usepackage{url}
\usepackage{verbatim}
\usepackage{graphicx}
\usepackage{cite}
\usepackage{multirow}
\usepackage{makecell}
\usepackage{booktabs}
\usepackage{ragged2e}

\begin{document}

\title{Developing Smart MAVs for Autonomous Inspection in GPS-denied Constructions}

\author{Yaoqiang Pan$^{1,*}$, Kewei Hu$^{1,*}$, Xiao Huang$^{2}$, Wei Ying$^{1}$, Xiaoxuan Xie$^{2}$,\\ Yue Ma$^{3}$, Naizhong Zhang$^{2,\#}$, and Hanwen Kang$^{1,\#}$
\thanks{$^{*}$ Equal Contribution} 
\thanks{$^{\#}$ Correspondence}
\thanks{$^{1} $ K.Hu, Y.Pan, W.Ying, and H.Kang are with the College of Engineering, South China Agriculture University, Guangzhou, China}
\thanks{$^{2} $ X.Huang, X.Xie, and N.Zhang are with the College of Civil Aviation, Nanjing University of Aeronautics and Astronautics, Nanjing, China}
\thanks{$^{3} $ Y.Ma is with the College of Mechatronics and Control Engineering, Shenzhen University, Shenzhen, China}}

\markboth{Journal of \LaTeX\ Class Files,~Vol.~14, No.~8, August~2021}%
{Shell \MakeLowercase{\textit{et al.}}: A Sample Article Using IEEEtran.cls for IEEE Journals}


\maketitle

\begin{abstract}
Smart Micro Aerial Vehicles (MAVs) have transformed infrastructure inspection by enabling efficient, high-resolution monitoring at various stages of construction, including hard-to-reach areas. Traditional manual operation of drones in GPS-denied environments, such as industrial facilities and infrastructure, is labour-intensive, tedious and prone to error. 
This study presents an innovative framework for smart MAV inspections in such complex and GPS-denied indoor environments. The framework features a hierarchical perception and planning system that identifies regions of interest and optimises task paths. It also presents an advanced MAV system with enhanced localisation and motion planning capabilities, integrated with Neural Reconstruction technology for comprehensive 3D reconstruction of building structures. 
The effectiveness of the framework was empirically validated in a 4,000 m² indoor infrastructure facility with an interior length of 80 metres, a width of 50 metres and a height of 7 metres. The main structure consists of columns and walls. Experimental results show that our MAV system performs exceptionally well in autonomous inspection tasks, achieving a 100\% success rate in generating and executing scan paths. 
Extensive experiments validate the manoeuvrability of our developed MAV, achieving a 100\% success rate in motion planning with a tracking error of less than 0.1 metres.
In addition, the enhanced reconstruction method using 3D Gaussian Splatting technology enables the generation of high-fidelity rendering models from the acquired data. 
Overall, our novel method represents a significant advancement in the use of robotics for infrastructure inspection.
\end{abstract}


\section{Introduction}
\IEEEPARstart{M}{icro} Aerial Vehicles (MAVs) have transformed infrastructure inspection by providing reliable and efficient methods for various applications, including Structural Health Monitoring (SHM) and other critical tasks\cite{pena2024uav}. Equipped with high-resolution cameras and a range of sensors, MAVs can monitor all phases of construction activities cost-effectively, from site preparation to project completion and ongoing daily monitoring\cite{wang2024rapid}. 
Their portability and flexibility allow for rapid access to hard-to-reach areas\cite{zhang2024developing}, transmitting high-resolution photographs and videos in real time \cite{pan2024pheno}. 
This capability enables construction management teams to perform on-demand reviews of construction status, compare plans with actual progress, inspect for quality control, and manage materials effectively\cite{yu2022uav}. 
Furthermore, MAVs facilitate faster and more efficient surveying, mapping, 3D modelling, and earthwork volume calculations. 
By enhancing accuracy and saving time, MAVs are instrumental in maintaining the safety and integrity of infrastructure, ensuring that all aspects of construction are meticulously monitored and managed\cite{zhang2024reactive}. 

Traditional methods of deploying MAVs for inspecting infrastructure in GPS-denied environments, such as underground sites or enclosed industrial facilities, heavily rely on manual processes\cite{stefan2023multi}. 
These environments often have complex layouts and obstacles that make remote control of MAVs particularly challenging. As a result, these inspections are time-consuming, labour-intensive, cumbersome, and prone to human error\cite{zhang2024developing}.
Automating these inspections using MAVs can significantly improve efficiency, accuracy, and safety. However, inspecting infrastructure in GPS-denied environments with MAVs presents significant challenges. The absence of reliable GPS signals complicates autonomous navigation and inspection tasks, requiring MAVs to rely on alternative localization, mapping, and obstacle avoidance methods\cite{gao2023uav}. 
Emerging techniques such as visual odometry, LiDAR, and other sensor-based approaches become critical in these contexts. 
The lack of GPS signals necessitates sophisticated algorithms and robust hardware to ensure accurate positioning and navigation\cite{tan2021automatic}. 
Additionally, the intricate layouts and numerous obstacles in these environments add another layer of complexity. 
MAVs must navigate through narrow passages, avoid obstacles, and adapt to changing environmental conditions while maintaining stability and precision in their inspections\cite{dianovsky2023electromagnetic}. 
This requires advanced motion planning, real-time situational awareness, and accurate control systems to achieve comprehensive and precise performance \cite{pan2024novel}.

This study presents a novel framework for performing autonomous inspection tasks in indoor infrastructure environments.
Firstly, we introduce a hierarchical environmental understanding and planning method that enables MAVs to automatically identify regions of interest requiring inspection. Based on this, we develop an instance-aware planning method to generate task paths based on perception and an exploration strategy to allow MAVs to complete tasks in complex and dynamic environments.
Secondly, we develop a smart MAV system that encompasses accurate indoor localization and obstacle-free motion planning capabilities. By optimizing hardware and software, concerning Size, Weight, and Power (SWaP), our MAVs can achieve onboard autonomous inspection tasks in infrastructure indoor scenes without any human intervention.
Overall, the contributions of this study are fourfold:
\begin{itemize}
    \item Introduce a novel hierarchical perception and planning framework to enable autonomous MAV inspection in complex indoor infrastructure environments.
    \item Develop an advanced MAV system capable of accomplishing autonomous inspection flying in complex indoor scenes without any human intervention.
    \item Integrate our novel framework with the Neural Reconstruction approach to achieve immersive and complete reconstruction of the inspection region.
    \item Evaluate the presented solution in an indoor infrastructure environment (over $4,000 m^{2}$), providing empirical evidence of its efficacy and reliability.
\end{itemize}

The remainder of this paper is organized as follows.
Section \ref{section: review} reviews related works.
Sections \ref{section: planning} and \ref{section: system} describe our methods.
Section \ref{section: experiment} evaluates the proposed methods and Section \ref{section: conclusion} summarizes the conclusion of the research.

\section{Related Works} \label{section: review}
\subsection{Review on MAVs in Infrastructure Inspection}
MAVs offer significant advantages in SHM due to their flexibility, mobility, cost-effectiveness, and comprehensive coverage. 
Equipped with advanced sensors and cameras, MAVs can acquire high-resolution image data critical for inspecting various infrastructures, including bridges, railroad tracks, underground mines, and vessels.
In bridge inspections, MAVs detect structural damage with visual sensors and reconstruct 3D models for permanent geometric records. 
This capability is demonstrated by Aliyari’s systematic hazard identification methodology \cite{aliyari2022hazards} applied to Norway’s Grimsøy bridge.
Li’s method \cite{li2023automatic}, utilizing UAVs and Faster R-CNN algorithms, enhances crack detection by optimizing imaging distance and stability, employing the DJI M210-RTK for high-quality image capture.
For railroad infrastructure, Mu’s ACSANet framework \cite{mu2023adaptive} significantly improves the detection accuracy of small defects in steel structures, while Cui’s SCYNet framework \cite{cui2023skip} excels in real-time inspection of high-speed railway noise barriers. 
Addressing corrosion risks, Yu’s deep learning-based method \cite{yu2023amcd} enables efficient visual inspections of steel structures. 
Additionally, Demkiv’s application \cite{demkiv2021application} of stereo thermal vision on drones effectively identifies overheated equipment in power lines.
MAVs also present major progress on inspection tasks in GPS-denied environments. 
Mansouri’s framework \cite{mansouri2019visual} utilizes CNN to detect tunnel crossings and junctions, highlighting MAVs' potential in subterranean inspections. 
Ortiz \cite{ortiz2016vision} discusses using MAVs with advanced vision systems and autonomous navigation to inspect ships for structural defects. 
Bonnin-Pascual \cite{bonnin2019reconfigurable} introduced a reconfigurable framework that transforms MAVs into versatile vessel inspection tools, allowing human surveyors to remotely control the MAVs while they autonomously handle safety tasks such as collision avoidance.
To improve the inspection performance, Cui and Dong et al. \cite{dong2024neural, cui20243d} applied the Neural Radiance Field (NeRF) in the 3D reconstruction of construction scenes for the inspection of large building structures.

\subsection{MAVs in GPS-denied Environments}
In recent years, the deployment of MAVs in GPS-enabled outdoor environments has increased significantly. 
However, their operation in GPS-denied environments remains challenging due to issues such as localization, navigation, environmental perception, and obstacle avoidance \cite{zhang2024developing}. 
These challenges are compounded by harsh conditions such as poor illumination, narrow passages, dirt, high humidity, and dust.
To address these challenges, numerous researchers have proposed reliable methods to enhance localization in GPS-denied settings. 
A common approach involves multi-sensor data fusion, utilizing sensors such as LIDAR, cameras, ultrasonic sensors, and IMUs to achieve accurate positioning.
Abraham Bachrach \cite{bachrach2011range} introduced a multi-level sensing and control hierarchy that aligns algorithm complexity with MAVs’ real-time needs, which was validated in indoor and urban canyon environments. 
Similarly, Kartik Mohta \cite{mohta2018fast} developed a system design and software architecture for autonomous MAV navigation, enabling swift and reliable target acquisition while avoiding obstacles in cluttered environments. 
Yingcai Bi \cite{bi2017robust} integrated a 2D laser scanner, camera, onboard computer, and flight controller to achieve state estimation and flight management in GPS-denied conditions.
Addressing the challenge of obstacle-free motion planning, Younes Al \cite{younes2021optimal} employed a Nonlinear Model Predictive Horizon (NMPH) method, allowing drones to navigate subterranean environments with smooth, collision-free paths that account for vehicle dynamics and real-time obstacles. 
Additionally, Shaekh Mohammad \cite{shithil2022robust} proposed a robust multi-sensor fusion-based method for navigating cluttered, dynamic, large-scale GPS-denied forest environments.

However, most drone systems that were developed for construction monitoring heavily rely on human remote control, making operations cumbersome, time-consuming, and sometimes dangerous, especially in indoor infrastructure or building scenes. Additionally, integrating a more expressive representation of the reconstruction model with MAV technology in construction inspections remains challenging.
\begin{figure*}
    \centering
    \includegraphics[width=1\linewidth]{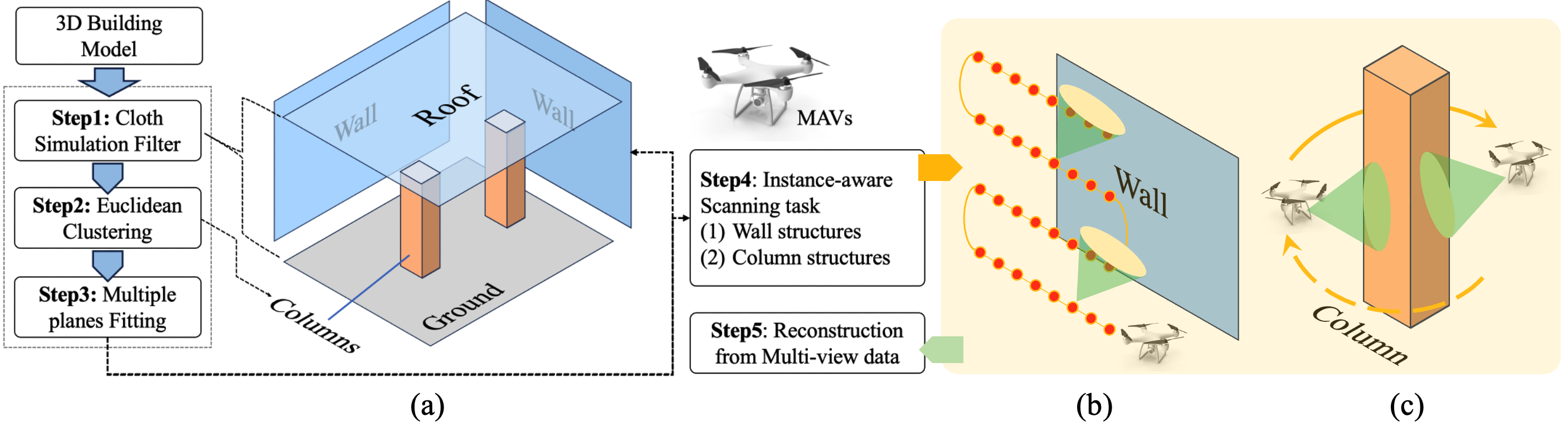}
    \caption{Framework of the building inspection MAVs system }
    \label{fig:framework}
\end{figure*}

\section{Hierarchy Planning Framework for SHM} \label{section: planning}
Our novel hierarchy planning framework is featured by its strong capability to understand the construction.
Overall, our framework includes three components: a semantic-level understanding algorithm, an instance-aware SHM scan-path planning method, and an execution strategy tailored for construction scenes, as detailed in Sections \ref{sec: semantic-process}, \ref{sec: instance plan}, and \ref{sec: SHM-exploration}, respectively,  as illustrated in Fig. \ref{fig:framework}

\subsection{Structur-aware Perception} \label{sec: semantic-process}
This section first describes the processing methods used to classify structures from the point cloud map $\mathcal{M}_{g}$, with the goal of representing the structures of interest as $\{\mathcal{O}\} \in \mathbb{R}^{N \times 3}$.

\subsubsection{Identification of Ground and Roof}
We identify the multi-level ground plane $\mathcal{O}_{ground} \in \{\mathcal{O}\}$ of the construction using the Cloth Simulation Filter (CSF) algorithm \cite{zhang2016easy}, by simulating a virtual cloth dropped onto an inverted point cloud. 
The virtual cloth is modelled as a grid of interconnected particles with mass, effectively separating ground points. 
The relationship between particle positions and forces is governed by Newton’s second law, as expressed by:
\begin{equation}
    m\frac{\partial X(t)}{\partial t^{2}}=F_{gravity}+F_{spring}+F_{damping}
\end{equation}
where $F_{\text{gravity}}$, $F_{\text{spring}}$, and $F_{\text{damping}}$ represent the forces due to gravity, and the spring-like connections between particles, and damping, respectively
Following this, the RANSAC algorithm is employed to fit a plane to $\mathcal{O}_{ground}$ and calculate the normal vector $N_g$ of the extracted ground plane. 
Further, the roof of the buildings $\mathcal{O}_{roof} \in \{\mathcal{O}\}$ can be extracted by computing its normal vector $N_r$, which satisfies the condition $N_r \cdot N_g = -1$.

\subsubsection{Identification of Column Structures}
Columns have a different cross-sectional area than walls, this feature can be used to classify them. 
A Euclidean clustering method \cite{sarle1991finding} is utilised to extract columns here. 
The process begins by constructing a Kd-tree in point cloud maps, facilitating rapid nearest neighbour searches. 
Points within a predefined distance threshold are iteratively clustered into $Q$, ensuring that each cluster contains points that are close to each other. 
This process continues until no additional points can be added to $Q$.
Specifically, columns $\mathcal{O} _{column}\in \{\mathcal{O}\}$ are identified based on their unique geometric properties within the clustered results of Euclidean clustering.

\subsubsection{Identification of Wall Structures}\label{section: wall_ex }
Wall structures often consist of multiple planes oriented in different directions. 
To address this, a plane fitting algorithm based on normal features is employed to extract each wall plane accurately. 
When fitting the plane $a x + by + cz + d = 0$, it is essential not only to minimize the distance from the plane to all points but also to ensure that the normal vector of the fitted plane $(a, b, c)$ is as consistent as possible with the normal vectors of the points in the point cloud, as expressed in Eq.\ref{EQ:distance}.

\begin{equation}\label{EQ:distance}
    \begin{split}
        \text{min} & \sum_i \left( \frac{|ax_i + by_i + cz_i + d|}{\sqrt{a^2 + b^2 + c^2}} \right)^2
    \end{split}
\end{equation}

\begin{equation}\label{EQ:distance}
    n_{xi} \cdot a + n_{yi} \cdot b + n_{zi} \cdot c = \sqrt{n_{xi}^2 + n_{yi}^2 + n_{zi}^2}
\end{equation}

where $a$, $b$, $c$, and $d$ are the parameters of the plane equation, where $(a, b, c)$ represents the normal vector of the plane and $d$ is the distance from the origin to the plane along its normal vector. 
The coordinates $(x_i, y_i, z_i)$ denote the points in the point cloud. 
$n_{xi}$, $n_{yi}$, and $n_{zi}$ are the components of the normal vector at point $i$ in the point cloud. 
By incorporating both distance minimization and normal vector consistency, the plane fitting can accurately identify all the $\mathcal{O} _{wall}\in \{\mathcal{O}\}$.

\subsection{Instance-aware SHM task Planning} \label{sec: instance plan}
Complete visual data acquisition of structures is crucial for SHM tasks. 
Therefore, associating the Field Of View (FOV) of cameras with the shooting distance is essential to secure inspection performance. 
The identified structure instances \(\mathcal{O}_{ground}, \mathcal{O}_{roof}, \mathcal{O}_{wall}, \mathcal{O}_{column} \in \{\mathcal{O}\}\) in Section \ref{sec: semantic-process} provide MAVs with location and geometric characteristics as prior knowledge for scan-path planning.  

\begin{figure}[ht]
    \centering
    \includegraphics[width=1\linewidth]{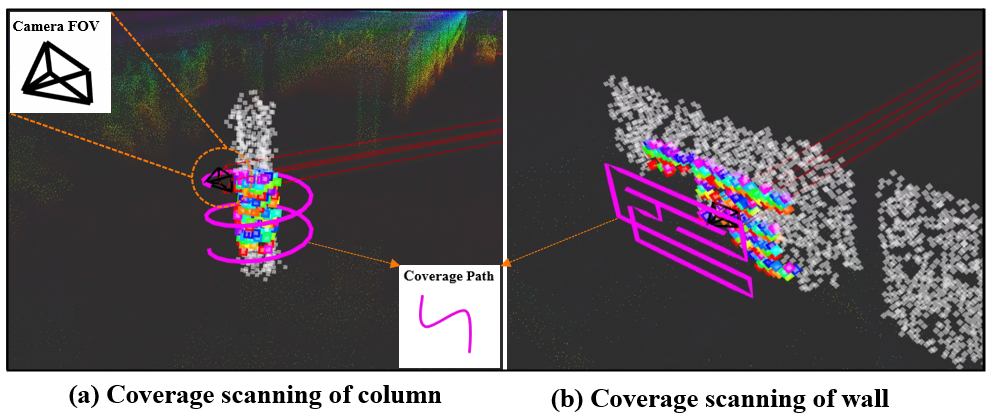}
    \caption{Visualization for MAV Coverage Scanning of Building structures}
    \label{fig:Covering Scanning}
\end{figure}

\subsubsection{MAV's FOV Calculation} 
The intrinsic parameters of the camera \((f_x, f_y, c_x, c_y)\) is first calibrated using the method described in \cite{zhang1999flexible}. 
Here, \(f_x\) and \(f_y\) represent the focal lengths in the x and y directions, respectively, and \(c_x\) and \(c_y\) are the coordinates of the principal point. 
The camera's field of view (FOV) is represented by \(\mathrm{FOV}_\mathrm{width}\) and \(\mathrm{FOV}_\mathrm{height}\), which are the horizontal and vertical extents of the FOV. 
These can be computed as follows:
\begin{equation}
 \left\{
      \begin{aligned}
    \mathrm{FOV}_\mathrm{width}=2\cdot D\cdot\tan\left({\theta_x}/{2}\right)\\
     \mathrm{FOV}_\mathrm{hight}=2\cdot D\cdot\tan\left({\theta_y}/{2}\right)
     \end{aligned}
     \right.
\end{equation}
where $\theta_x$ ,$\theta_y$  are horizontal angle and vertical angle are computed by $\theta_x=2\cdot\arctan\left({c_x}/{f_x}\right)$, $\theta_y=2\cdot\arctan\left({c_y}/{f_y}\right)$.


\subsubsection{Scan-Path generation} 
The $\mathcal{O}_{ground}, \mathcal{O}_{roof}$ in map limit the safe flight height from $h_{max}$ to $h_{min}$ , while  $\mathcal{O}_{column},\mathcal{O}_{wall}$ are structure of  inspection interest. 
Two different strategies are respectively designed to generate the scan path for SHM tasks.

\textbf{(a) Column Structures}: For each \(\mathcal{O}_{column}\) instance, we utilize a spiral path generation strategy that considers the camera's FOV to ensure complete scan coverage. 
The radius \(r_{\mathrm{spiral}}\) is determined by the sum of the \(\mathcal{O}_{column}\) instance's cross-sectional radius \(r\) and the shooting distance \(D\). 
Given that the height of the spiral ranges from \(h_{max}\) to \(h_{min}\), the final spiral scan path is generated using the following equation:

\begin{equation}
\left\{
\begin{aligned}
x_i &= r_{\mathrm{spiral}} \cos(\theta_i) \\
y_i &= r_{\mathrm{spiral}} \sin(\theta_i) \\
z_i &= \frac{h_{max} - h_{min}}{2\pi n} \cdot \theta_i + h_{min}
\end{aligned}
\right.
\end{equation}
where the number of turns \( n \) is computed as \(\left\lceil \frac{h_{max} - h_{min}}{\mathrm{FOV}_{\mathrm{height}}} \right\rceil\). 
The angle \(\theta_i\) for the \(i\)-th point on the spiral path is given by \(2 \pi i / m\), where \(m\) is the total number of waypoints in one complete turn of the spiral. 
The value of \(m\) is determined by dividing the circumference of the \(\mathcal{O}_{column}\) instance (i.e., \(2 \pi r\)) by the \(\mathrm{FOV}_{\mathrm{width}}\). 
Here, \(i\) ranges from \(0\) to \(m \cdot n\).


\textbf{(b) Wall Structures}: To ensure that the inspection path covers the full area of each \(\mathcal{O}_{wall}\) instance, we first project every 3D point of the wall instance onto a plane at the shooting distance \(D\) based on the plane's normal direction. 
Image processing techniques are then used to find the minimum bounding rectangle, denoted as \(\mathcal{M}_{b}\), of this region on the 2D plane. 
Subsequently, a Complete Coverage Path Planning (CCPP) algorithm, described in Algorithm \ref{alg:ccpp}, is utilized to generate the scanning path on the wall plane. 

\begin{algorithm}
\caption{Complete Coverage Path Planning (CCPP)}
\label{alg:ccpp}

\textbf{Input:} \( \mathcal{M}_b \) (bounding map), Starting Position \( P_o \)

\textbf{Output:} \( \textbf{path\_Vec} \)

\begin{algorithmic}[1]
\STATE Initialize \( \textbf{path\_Vec} \) as empty queue
\STATE \( \text{current\_position} \leftarrow P_o \)

\STATE \textbf{Initialize\_map}(\( \mathcal{M}_b \)):
    
\STATE \textbf{Cover\_planning}():
    \WHILE{not complete\_coverage}
        \STATE \( \text{neighbors} \leftarrow \text{evaluate\_neighbors}(\text{current\_position}, \mathcal{M}_b) \)
        \STATE \( \text{next\_direction} \leftarrow \text{max\_state\_value\_direction}(\text{neighbors}) \)
        \STATE \( \text{move\_robot}(\text{current\_position}, \text{next\_direction}) \)
        \STATE \( \textbf{path\_Vec.append}(\text{current\_position}) \)
    \ENDWHILE

\end{algorithmic}
\end{algorithm}

\textbf{Initialize\_map}: 
The map is partitioned into grid cells. 
Each cell is marked as covered if traversed by the robot. 
Cells with obstacles are assigned a value of -1000, while free cells receive a value of \({50}/{j}\), where \(j\) is the column number. 


\textbf{Cover\_planning}: the robot's current position serves as the starts \(P_o\) for scan path planning. 
Iteratively, the algorithm evaluates the state values of adjacent cells in eight directions (0°, 45°, 90°, 135°, 180°, 225°, 270°, 315°). 
The direction with the highest state value is chosen as the goal. 
The robot advances one grid cell in this direction, updates its position and appends it to the path \textbf{path\_Vec}. 
This process continues until the complete covering is reached.


\subsection{Inspection Execution by Exploration} \label{sec: SHM-exploration}
To avoid unsafe flying caused by unknown changes and the complex nature of the construction scenes, an exploration strategy is introduced and implemented before executing SHM tasks, as described in Alg.\ref{alg:expl_strategy}.

\begin{algorithm}
\caption{Exploration-based SHM Execution}
\label{alg:expl_strategy}
\quad\textbf{Input:} Structure Instance List $L_{\mathcal{O}}$, Scanning Path List $L_{\mathcal{T}}$

\quad \textbf{Output:} Collision-free Scanning Path List $L_{\mathcal{T}_f}$

\begin{algorithmic}[1]
\FOR{ $\mathcal{O}_i$ in $L_\mathcal{O}$}
\STATE $\mathcal{M}_i,\alpha_i \leftarrow \text{\textbf{ResetExploreMap}}()$\;
\WHILE{$\alpha_i<\tau$}
    \STATE ${G} \leftarrow \text{\textbf{GenExplorationGoals}}(\mathcal{O}_i)$\;
    \STATE  \textbf{Plan$\&$MovetoGoal}($G$)
    \STATE  $\mathcal{M}_i,\alpha_i  \leftarrow \textbf{UpdateTaskmap}(C_{t_0:t_0+T})$
\ENDWHILE
\STATE Update $L_{\mathcal{T}_f}(i)$ $\leftarrow$ \textbf{Check$\&$Replan}($\mathcal{M}_i$, $L_{\mathcal{T}}(i)$)
\STATE \textbf{Execution}($L_{\mathcal{T}_f}(i)$)
\ENDFOR
\end{algorithmic}
\end{algorithm} 

 The algorithm takes as input a list of structure instances $L_{\mathcal{O}}$ and a list of scanning paths $L_{\mathcal{T}}$. 
 For each structure instance $\mathcal{O}_i$ in $L_{\mathcal{O}}$, it initializes a local task map $\mathcal{M}_i$ and an exploration rate $\alpha_i$ (0 at the start) using the \textbf{ResetExploreMap} function. 
 While $\alpha_i$ is below a specified threshold $\tau$, exploration goals $G$ are generated using 
 
 \begin{figure}[h]
    \centering
    \includegraphics[width=0.93\linewidth]{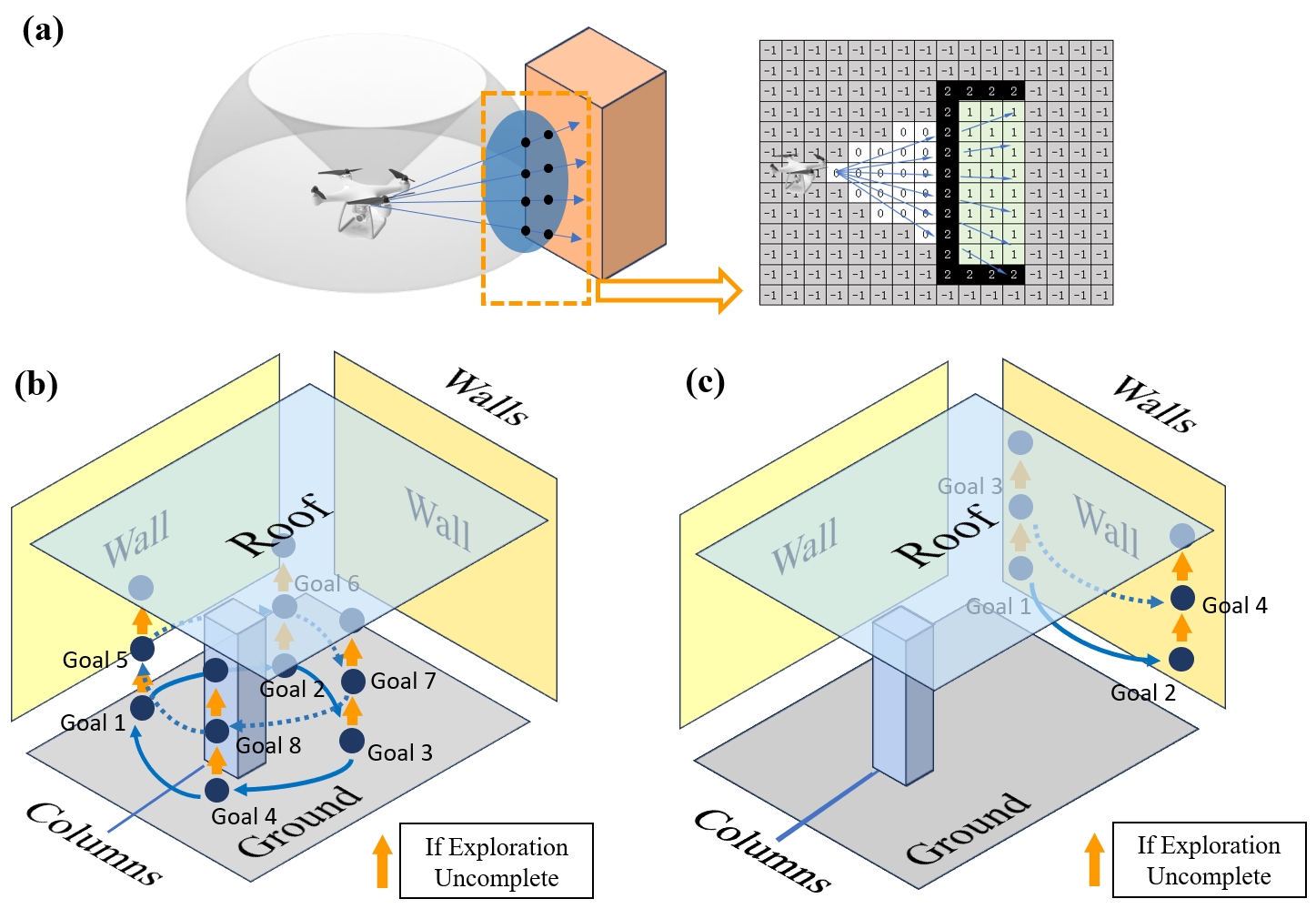}
    \caption{Exploration Procedures Demonstration (a) Obstacle Raycasting Modeling Demonstration (b) Exploration Procedures of Column Structures (c) Exploration Procedures of Wall Structures}
    \label{fig:exploration}
\end{figure}
 
 \textbf{GenExplorationGoals}, and the system plans and moves to the goals with \textbf{Plan\&MovetoGoal} (see Sec \ref{sec: mvs-planner}). 
 The task map $\mathcal{M}_i$ and progress $\alpha_i$ are updated based on the LiDAR point cloud frames $C_{t_0:t_0+T}$. 
 Once the exploration is sufficient, the algorithm updates the collision-free scanning path $L_{\mathcal{T}_f}(i)$ by replanning paths using \textbf{Check\&Replan}. 
Finally, the updated path is executed for each structure.

\textbf{GenExplorationGoals}: Different instance objects $\mathcal{O}_i$ generate specific exploration points, as depicted in Fig. \ref{fig:exploration}. 
For column objects, MAVs will choose the exploration points that are evenly spaced near the columns' structure centre. 
For wall objects, MAVs will search the space on the free side of the wall surface. 
If the exploration rate is below the threshold, subsequent attempts will adjust the flying height of exploration points until the task requirements are reached.

\textbf{UpdateTaskMap}: Using the principle of raycasting, voxel states in the local task map are determined from the LiDAR scan. 
Voxels along the path to a LiDAR point are considered free space and set to 0, while the voxel at the obstacle point is assigned a value of 1. 


\section{Development of Smart MAVs} \label{section: system}
This section introduces the MAVs, which include two fundamental components, MAVs localization and motion planning, as detailed in Sections \ref{sec: mvs-localization} and \ref{sec: mvs-planner}, respectively.
\begin{figure}[ht]
    \centering
    \includegraphics[width=0.85\linewidth]{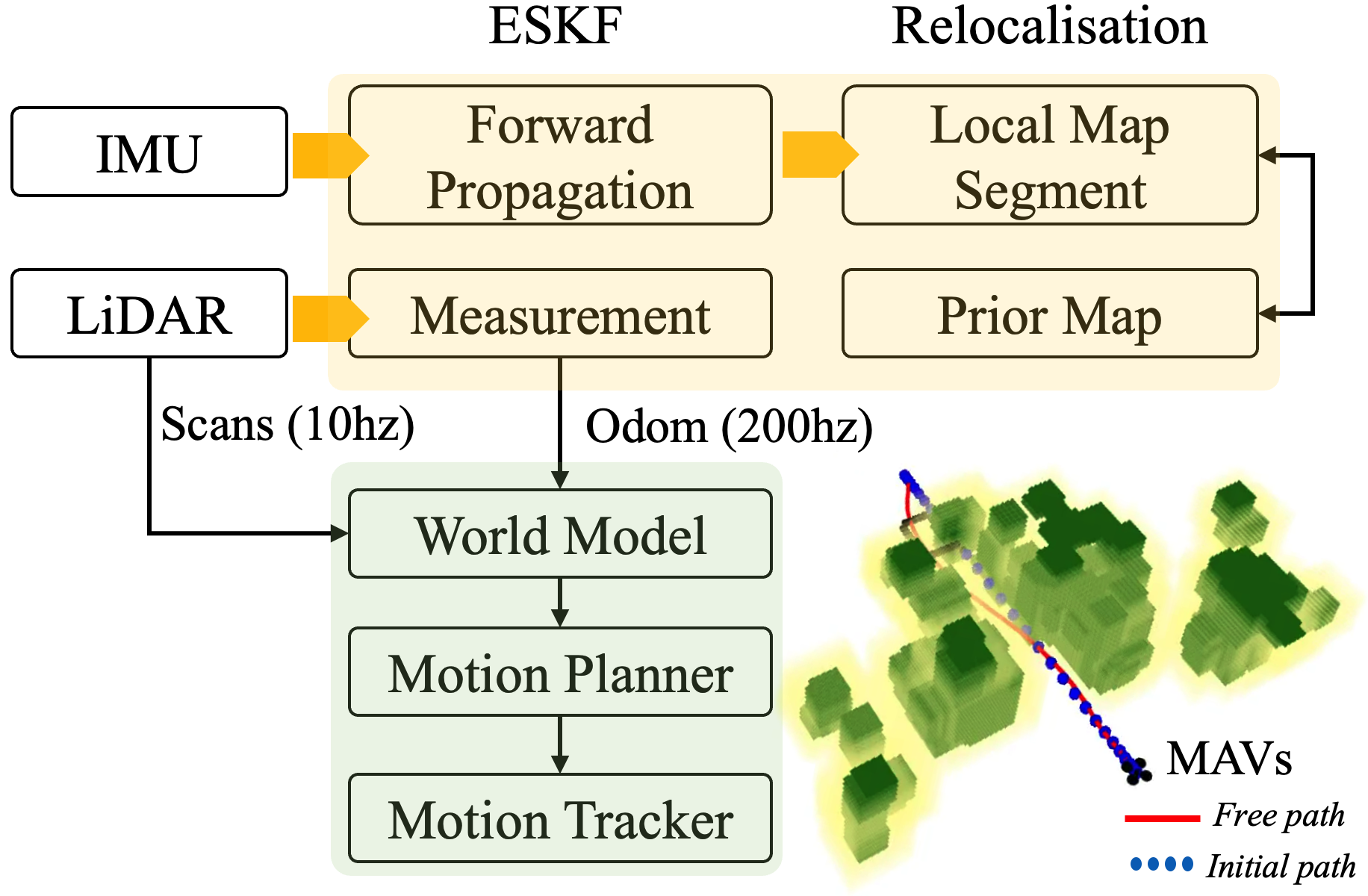}
    \caption{Semantic instance extracting procedure }
    \label{fig:MAVs}
\end{figure}

\subsection{MAVs Localisation in GPS-denied Building Indoor} \label{sec: mvs-localization}
An Error-Sate Kalman Filter (ESKF) is adopted for MAVs state estimation by using the LiDAR and IMU measurements. The system state $x$, error-state $\delta x$, IMU measurement input $u$, and measurement noise $w$ can be expressed as:
\begin{equation}
\begin{split}
    &x=\{^{G}R_{M}\ ^{G}P_{M}\ ^{G}v_{M}\ b_{\omega}\ b_{a}\ b_{g}\} \\
    &\delta x = \{\delta \theta\ \delta P\ \delta v\ \delta b_{\omega}\ \delta b_{a}\ \delta b_{g}\} \\
    &u=\{\omega_{M}\ a_{M}\} \qquad w=\{n_w\ n_a\ n_{b_{\omega}}\ n_{b_a} \}
\end{split}
\end{equation}
where $G$ donates the global frame (we use the first IMU frame aligned with gravity direction), $M$ denotes the body frame of the MAVs. 
$^{G}R_{M} \in SO(3)$, $^{G}P_{M} \in R^{3}$, and $^{G}v_{M}$ are the position, attitude, and velocity in the global frame.
$\omega_{M}$ and $a_{M}$ are the angular velocity and linear acceleration from IMU measurements.
$b_{\omega}$, $b_{a}$, and $b_{g}$ are the bias. $n_w$ and $n_a$ are the noise of the IMU measurements, while $n_{b_{\omega}}$ and $n_{b_a}$ are the random walk noise of $b_{\omega}$ and $b_{a}$, respectively

\subsubsection{Forward Propagation}
We use $x$ and $\hat{x}$ to donate the nominal and true states and $x = \hat{x}+\delta x$. 
Once receiving an IMU measurement, the state propagation can be calculated by:
\begin{equation}
\begin{split}
    &\hat{x}_{k}=x_{k-1} \oplus [\Delta t \cdot f(x_{k-1}, u_k, w_k)] \\
    &\delta \hat{x}_{k} = F_{x}\delta x_{k-1}+F_{w} w_{k} \\
    &\hat{P}_{k}=F_x P_{k-1} F_x^{T} + F_w Q_{k} F_w^{T}
\end{split}
\end{equation}
where $P_{k-1}$ is the propagated covariance of the $\delta x$ at index $k-1$ and $Q_{k}$ is the covariance of the noise $w$ and index of $k$.
where $f(x_{k-1}, u_k, w_k)$ are the discrete kinematic model of IMU propagation [], $F_x$, and $F_i$ are defined as:
\begin{equation}
\begin{split}
        F_x &= \left. \frac{\partial f(\cdot)}{\partial \delta x} \right|_{\mathbf{x_{k-1}}, \mathbf{u}_m}
        F_w = \left. \frac{\partial f(\cdot)}{\partial \delta w_{k}} \right|_{\mathbf{x_{k-1}}, \mathbf{u}_m}
\end{split}
\end{equation}

\subsubsection{Measurement and Update}
We utilise a GICP algorithm to register LiDAR scan.
When a new LiDAR scan is reached, for each point $p_{i}$, its correspondence is found by searching the nearest $q_{m_{i}}$ in the grid map.
The cost function $D_i$ between the correspondence matches is defined by:
\begin{equation}
    D_{i} = [( q_{m_{i}} - \mathbf{T}p_{i})^T \frac{w(C + \lambda I)^{-1}}{\| (C + \lambda I)^{-1} \|_F} ( q_{m_{i}} - \mathbf{T}p_{i})]
\end{equation}
where $w$, $I$, and $\lambda$ are the weight, identity matrix, and constant, respectively, and $\|\cdot\|_{F}$ indicates the Frobenius norm. $C$ is the distribution covariance, indicating the planner feature in grid $q$. The observation $y_{k}$ can be calculated by optimizing:
\begin{equation}
    y_{k} = \mathrm{arg} \min_{\mathbf{T}} \sum_{i=0}^{N} \| D_{i} \|
\end{equation}
Then, we follow the iterated update mechanism of the ESKF, by evaluating the following:
\begin{equation}
    \begin{split}
    &\begin{cases}
    \delta \hat{x}_{k}=K_{k}(y_{k}-h(\hat{x}^{\kappa-1}_{k}+ \delta \hat{x}_{k})) \\
    \hat{x}^{\kappa}_{k} = \hat{x}^{\kappa-1}_{k} + \delta \hat{x}_{k} \\
    \end{cases} \\
    & \mathrm{repeat\ until} \ \| \delta \hat{x}_{k} \|<\epsilon
    \end{split}
\end{equation}
where $K_k$ is the Kalman gain, $h$ is the observation model, and we set $ \epsilon =0.1$ to indicate the convergence of the update. 
We recommend readers for the study [] for detailed derivation.

\subsubsection{Global Relocalisation}
Given a local map $M_{mvs}$ that is reconstructed in real-time, a dense point cloud registration method [] is used to find its pose $T_{ex}$ in prior map $M_{prior}$ by solving the function:
\begin{equation}
    \mathbf{T}_{ex} = \mathrm{arg} \min_{\mathbf{T}_{ex}}  \| M_{mvs} - \mathbf{T}_{ex} M_{prior}\|
\end{equation}
this pose $T_{ex}$ will be updated every five seconds, enabling MAVs to know its transient state in the prior-built map.

\subsection{MAVs Motion Planning and Control} \label{sec: mvs-planner}
\subsubsection{World Model}
We utilised the grid map to represent the obstacles. 
Different occupancy states are represented by distinct values (Fig.\ref{fig:MAVs}): 0 indicates a free space, 1 presents an obstacle (green), and 2 represents the expansion areas near obstacles (yellow). 
The hash table is used and the indexes of points are computed by:

\begin{equation}
\begin{aligned}
\text{L} &= \begin{bmatrix} L_x & L_y & L_z \end{bmatrix}^\top \\
        &= \begin{bmatrix} \mathrm{floor}(\frac{p_x}{r}) & \mathrm{floor}(\frac{p_y}{r}) & \mathrm{floor}(\frac{p_z}{R}) \end{bmatrix}^\top \\
hash\_Index &= L_x + L_y \cdot n_x + L_z \cdot n_x \cdot n_y
\end{aligned}
\end{equation}
where $p_x$, $p_y$, $p_z \in \mathbb{R}$ are the position of LiDAR point. 
The function $\mathrm{floor}(\cdot)$ computes the largest integer value not greater than its argument. 
$r$ is the map resolution and is set as 0.1$m$. 
$n_x$ and $n_y$ are the prime numbers for computing the hash index, respectively, with $n_x=73856093$ and $n_y= 83492791$.

\subsubsection{Motion Planner}
The instance-aware planning module generates a scan path $\Phi(t) = \{Q_0, Q_1, \cdots, Q_n \ | \ Q_i \in \mathbb{R}^3 \}$ that does not account for collisions, where \(Q_i\) represents a control point on the scan path $\Phi(t)$. 
According to the uniform B-spline property, there is a constant time interval \(\Delta t\) between consecutive control points \(Q_i\) and \(Q_{i+1}\). 
Consequently, the dynamic properties of the scan path, including velocity \(v_i\), acceleration \(a_i\), and jerk \(j_i\) at control point \(Q_i\), can be computed as follows:
\begin{equation}
\begin{matrix}
v_{i}=\frac{Q_{i+1}-Q_{i}}{\triangle t} \  \ 
a_{i}=\frac{v_{i+1}-v_{i}}{\triangle t} \  \ 
j_{i}=\frac{a_{i+1}-a_{i}}{\triangle t}
\end{matrix}
\label{eq: v-a-j}
\end{equation}
For the control points $Q_i \rightarrow Q_m$ in the obstacle, the RRT$^*$ is used to find a collision-free path $p_i \rightarrow p_m$ in the free voxel. 
The obstacle-free path then can be obtained by replacing $Q_i$ with $p_{i}$, and the distance between the new scan path to the obstacle can be defined as:
\begin{equation}
    d_{i}=\|(Q_i-p_{i})\cdot v_{i}\|
    \label{eq: collision avoidance}
\end{equation}
We then compute a kinematic-feasible and collision-free path, by means of optimizing the $Q_i$ on $\Phi$. the cost function is:
\begin{equation}
    \min_{\Phi}\ J=\lambda_c \sum^{n}_{i} J_c(i)+ \lambda_sJ_s+\lambda_dJ_d
\label{eq: optimization}
\end{equation}
where $J_c$ is collision term, $J_s$ is dynamic smoothness term, and $J_d$ is dynamic feasibility term, $\lambda_c$, $\lambda_s$, $\lambda_d$ are weights of each term, respectively.

For collision avoidance, we set a minimum safe distance $S_{f}$ and ensure all the control points satisfy $d_{i} > S_f$. 
The collision term $J_{c}(i)$ of each control points in Eq.\ref{eq: optimization} is defined as:
\begin{equation}
\begin{aligned}
J_{c}(i)& =\begin{cases}0&(d_{i}>1.5 S_f)\\
                        3 (S_f-d_{i}) &(S_f<d_{i}\leq 1.5 S_f)\\
                        (S_f-d_{i})^3&(d_{i}\leq S_f)
            \end{cases}
\end{aligned}
\end{equation}
To secure the dynamic smoothness, we examine the squares of the acceleration and jerk at each control point. 
This smoothness term in Eq.\ref{eq: optimization} is therefore defined as:
\begin{equation}
\begin{aligned}
J_s=\sum_{i=1}^{n-1}\|a_i\|^2+\sum_{i=1}^{n-2}\|j_i\|^2
\end{aligned}
\end{equation}
To guarantee the motion feasibility, we further evaluate the dynamic profiles at all control points, ensuring they adhere to the imposed constraint.
The feasibility term is defined as:
\begin{equation}
\begin{aligned}
J_d=w_v\sum_{i=1}^{n}B(v_i)+w_a\sum_{i=1}^{n-1}B(a_i)+w_j\sum_{i=1}^{n-2}B(j_i)
\end{aligned}
\end{equation}
where $w_v, w_a,$ and $w_j$ are the weights of different higher-order derivatives. 
The $B(\cdot)$ is defined as:

\begin{equation}
\begin{aligned}
B(c_r)=\begin{cases}0&( c_m\le c_r\le c_m)\\(c_r-c_m)^3&(c_m<\|c_r\|<2c_m)\\c_r^2&(\|c_r\|\ge 2c_m)\end{cases}
\end{aligned}
\end{equation}
where $c_r\in\{v_i,a_i,j_i\}$ and $c_{m}$ are the allowed maximum dynamic profiles in velocity, acceleration, and jerk.

\subsubsection{Motion Tracker} 
The motion planner computes a path $\Phi(t)$, passing the goal state $\{x^t_{goal}, \dot{x}^t_{goal}, \ddot{x}^t_{goal} \}$, which includes the desired position, velocity, and acceleration. 
The required $\ddot{x}_c = \{ \ddot{x}^{\text{x}}_{c}, \ddot{x}^{\text{y}}_{c}, \ddot{x}^{\text{z}}_{c} \}$ to reach goal is computed by:
\begin{equation}
\begin{aligned}
    e_{p}&=x^t_{goal}-x^{t}_{odom}\\
    e_{v}&=\dot{x}^t_{goal}-\dot{x}^{t}_{odom}\\
    \ddot{x}_{c}&=\ddot{x}^t_{goal}+K_{d}e_{v}+K_{p}e_{p}
\end{aligned}
\end{equation}
where $K_d$, $K_p$ are the coefficient of proportional and differential error, $x^{t}_{odom}$, $\dot{x}^{t}_{odom}$ are the estimated position and velocity from odometry, respectively. 
Then, the attitude of the MAVs $q_{ctrl} = \{ \theta, \phi, \psi \}$ and thrust $u$ are computed by solving:
\begin{equation}
    \begin{aligned}
        \ddot{x}^{\text{x}}_{c}&=g(\theta cos\psi+\phi sin\psi) \\
        \ddot{x}^{\text{y}}_{c}&=g(\theta sin\psi-\phi cos\psi) \\
        \ddot{x}^{\text{z}}_{c}&=-g+\frac{u}{m}
    \end{aligned}
\end{equation}
where $\theta$, $\phi$, $\psi$ are the pitch, roll, and yaw, $g$ is the gravity and $m$ is the mass of the MAVs, respectively. Finally, the $q_{ctrl}$ and $u$ are sent to the Flight Centre Unit (FCU) for execution.


\section{Case Study in Subterranean Facility} \label{section: experiment}
\subsection{Description of Case Environments}
\begin{figure}[ht]
    \centering
    \includegraphics[width=1\linewidth]{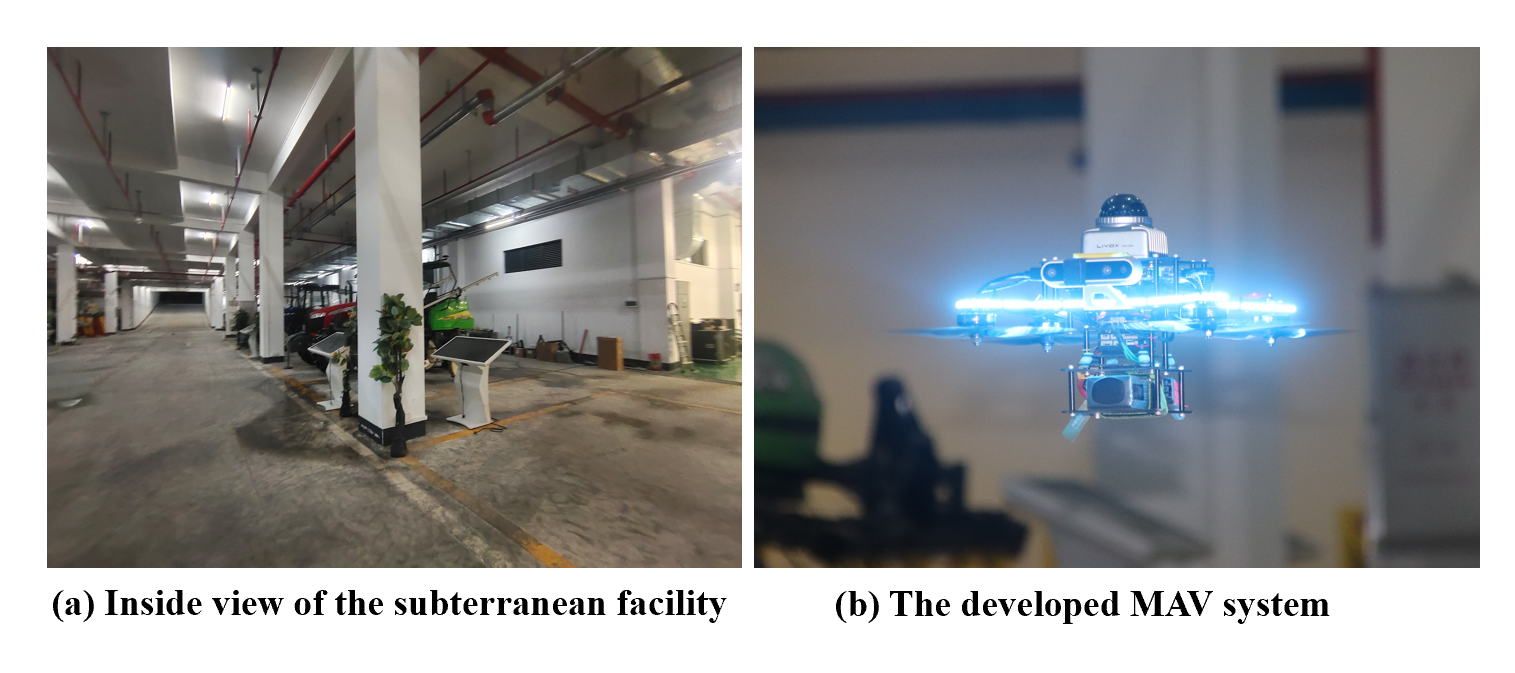}
    \caption{Case description of the subterranean facility}
    \label{fig:case-des}
\end{figure}

The developed system is evaluated in a subterranean facility at South China Agriculture University, Guangzhou, China. This facility has an interior length of 80m, a width of 50m, and a height of 7m, with the main structure consisting of columns and walls, as shown in Fig.\ref{fig:case-des}. In this study, we employed the developed MAV system to conduct automatic inspections within this facility, performing routine image data collection for normal monitoring and high-fidelity 3D reconstruction for areas requiring special inspection.
In this case, we will introduce the experimental setup and evaluation metrics to comprehensively assess the MAV system's performance in inspection tasks. The evaluation will include the quality of the image data, the effectiveness of the 3D reconstructions, and an analysis of the performance of the path planning and navigation system.

\noindent According to China's "Urban Underground Comprehensive Utility Tunnel Operation and Maintenance and Safety Technical Standard" (GB 51354-2019), regular monitoring of the main structures of buildings is required. This includes normal monitoring, which encompasses on-site inspections and remote video inspections, as well as special monitoring at specified times and locations to target potentially problematic areas.
\begin{figure*}[ht]
    \centering
    \includegraphics[width=1\linewidth]{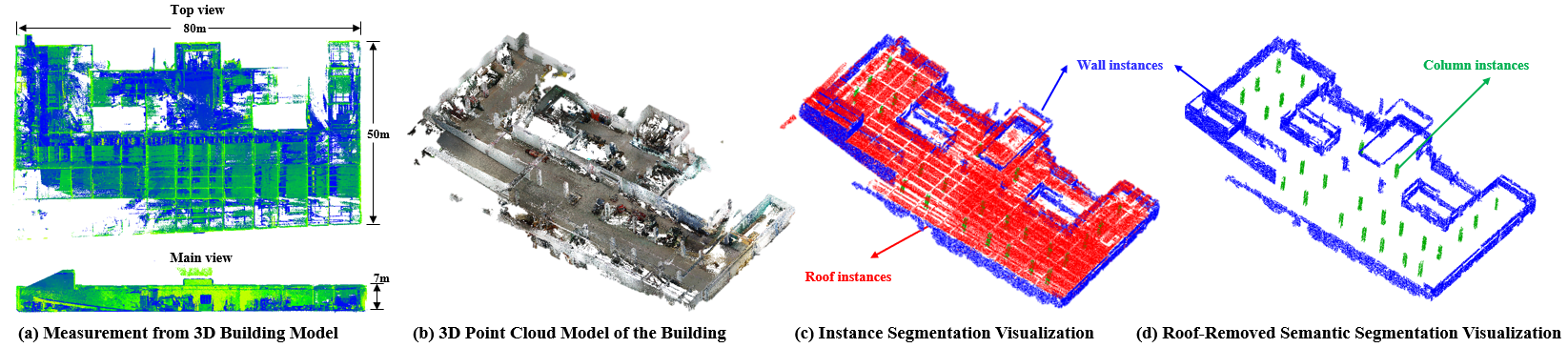}
    \caption{The results of the structural instance identification}
    \label{fig:identification}
\end{figure*}

\subsection{Software and Hardware Implementation}
The MAVs' odometry and relocalization module is adapted from the open-source Fast-LIO2 \cite{xu2022fast} and VGICP \cite{koide2021voxelized}, with enhancements to accuracy and computational speed tailored for construction environments. 
The MAVs' planning and control systems are implemented utilising a customized version of Ego-Planner \cite{zhou2020ego} and PX4 controller. 
This offers a more stable environmental modelling ability and a regulated motion optimization process. 
The software is built on the Robot Operating System (ROS) Noetic, running on Ubuntu 20.04.
Our MAVs (See Fig. \ref{fig:case-des} (b)) feature a lightweight design with four SUNNYSKY X-2216 II KV1280 brushless DC motors controlled by XF-35A ESCs. 
The Pixhack-6c FCU communicates via Dshot600 with ESCs and UART with the onboard computer. 
The sensor kit includes a LIVOX-MID360 LiDAR and RealSense D435 depth camera. 

\begin{figure}[ht]
    \centering
    \includegraphics[width=0.83\linewidth]{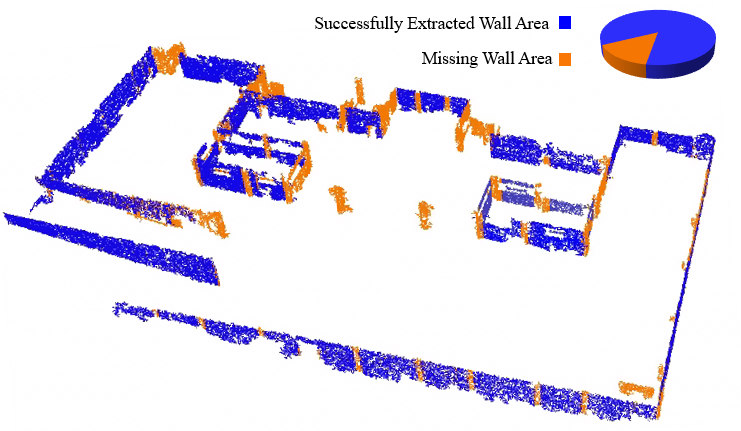}
    \caption{The result of multiple wall planes extraction}
    \label{fig:wall-result}
\end{figure}

\subsection{Experiment 1: Evaluation of Inspections Performance}
First, the point cloud model of the construction is initially built using a LiDAR-based SLAM system, with the mapping process performed by LIO-SAM \cite{shan2020lio}. 
Then, according to the method proposed in Sec.\ref{sec: semantic-process}, the wall and column structures in the point cloud model are identified and extracted. 
The quality of the visual data automatically collected by the UAV is evaluated using a no-reference method. 
Based on the evaluation results, we remove images with motion blur or distortion. Finally, we used the selected high-quality images to perform 3D reconstructions of the building structures.

\subsubsection{Identification of Structural Instances} \label{section: dq}
The results of our structure identification method are presented in Fig.\ref{fig:identification} (c) and (d) (refer to Section \ref{sec: semantic-process}). 
In these figures, the roof and ground instances are highlighted in red. 
Additionally, Fig.\ref{fig:identification}(d) provides a more detailed view of the internal segmentation results following the removal of roof obstructions, with wall instances depicted in blue and columns marked in green.
To assess the accuracy of our identification results, we compared them against the building structure ground truth obtained through manual labelling. 
The accuracy was quantified using the F1 score \cite{saito2015precision}.

\begin{figure}[ht]
    \centering
    \includegraphics[width=1\linewidth]{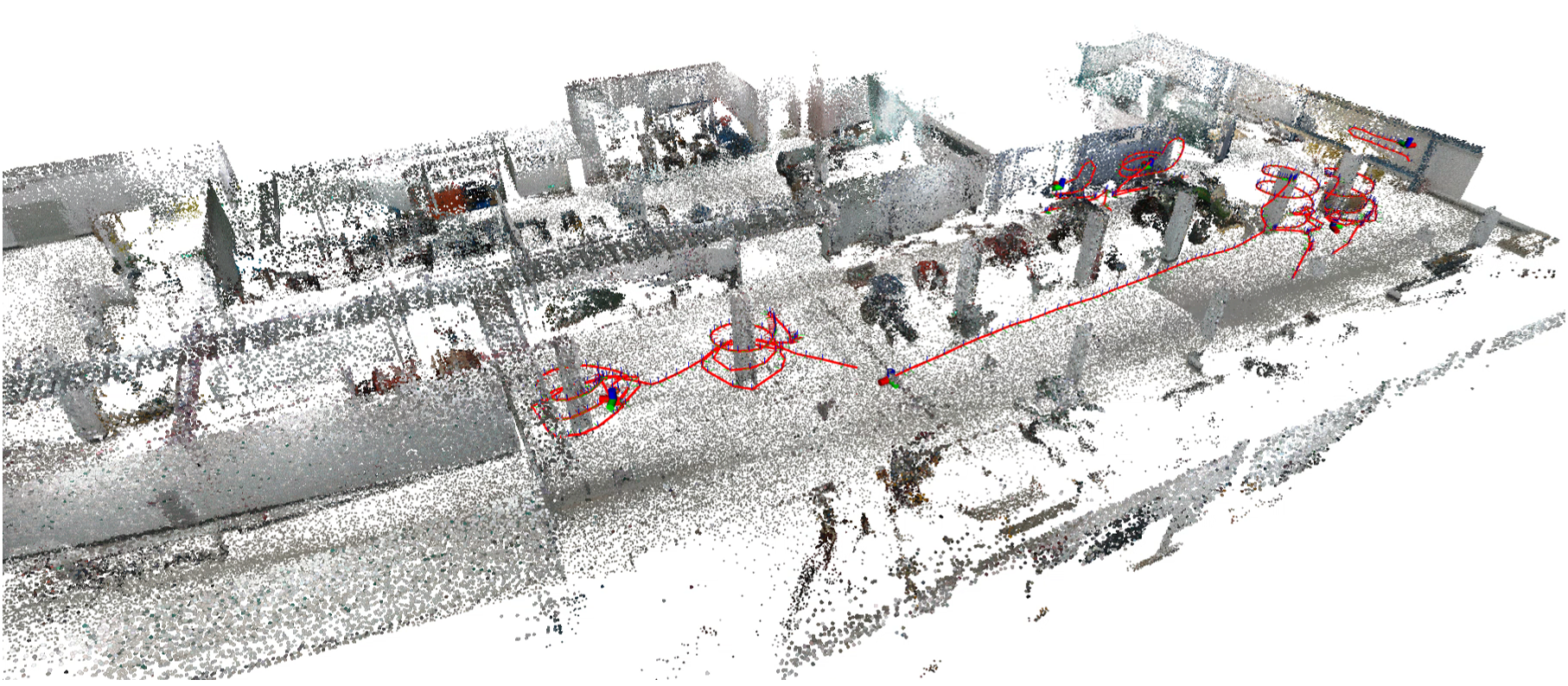}
    \caption{Overview of MAVs' path in two scan flights (red path)}
    \label{fig:total_flight}
\end{figure}

In total, there are 27 columns in the facility, and our method successfully segmented 25 of them, resulting in an F1 score of approximately 0.962 in this case study. For structural components like the roof and ground, which are integral parts of the building, we performed a qualitative analysis of the segmentation results. As observed in Fig.\ref{fig:identification} (c) and (d), the roof and ground can be completely removed from the model, confirming the accuracy of our segmentation process.
Furthermore, using the method described in Sec.\ref{section: wall_ex}, we performed a multi-plane extraction of the wall structures. The results of this process are shown in Fig.\ref{fig:wall-result}, where the blue areas represent successfully extracted wall information and the orange areas indicate regions omitted by the multi-plane extraction algorithm. By comparing these sections, we can determine the proportion of the wall instance extraction relative to the entire wall structure. 
In this case study, we utilized CloudCompare software to calculate the number of points representing successfully extracted wall instances, corresponding to the volume of the wall structure. By comparing this extracted volume to the actual total volume of the wall, we determined that our method successfully extracted 85\% of the total wall structure.

\begin{figure*}
    \centering
    \includegraphics[width=1\linewidth]{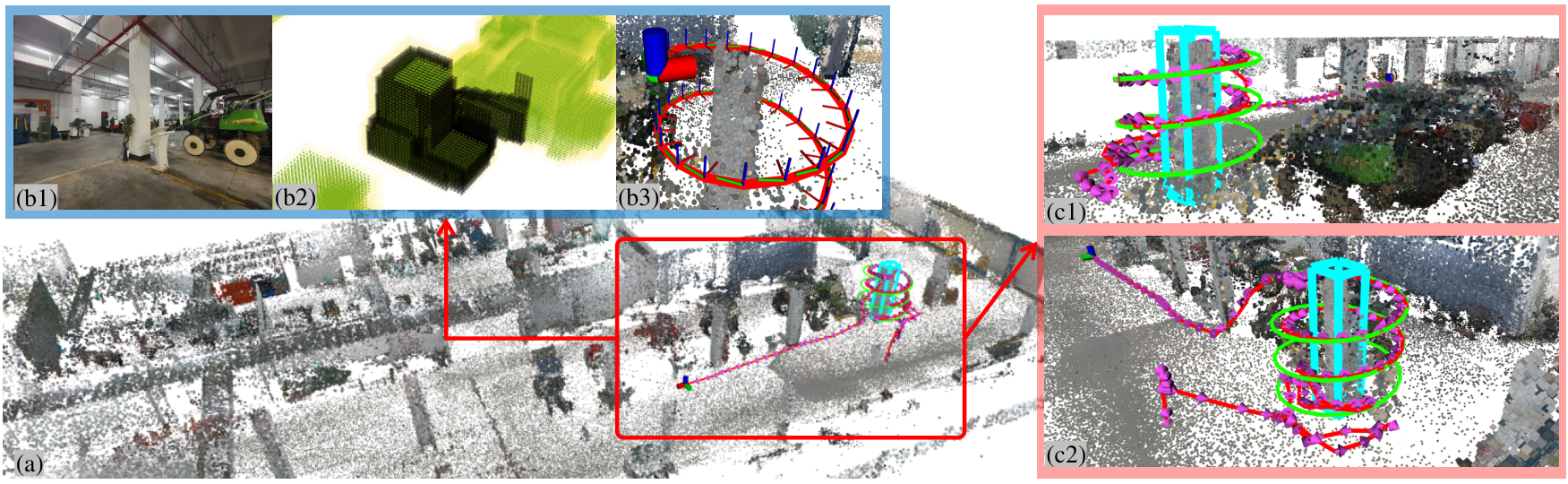}
    \caption{(a) Full Process Demonstration of the Scanning Task (b1) Real-World Scenario of the Scanning Task for Column Structures (b2) Obstacle Map acquired by exploration actions (b3) MAVs Executing Scanning Task Trajectory considering obstacles (c1), (c2) Different Perspectives of the Executed Trajectory (Green: Reference Path, Purple Arrows: Actual Orientation of MAVs)}
    \label{fig: task_process}
\end{figure*}

\subsubsection{Automatic Scanning by MAVs} 
In this section, an MAV was tasked with a specific scanning mission to evaluate its autonomous scanning performance during the inspection process (refer to Fig. \ref{fig:total_flight}). 
The execution sequence of this task is detailed in Fig. \ref{fig: task_process}. 
Fig. \ref{fig: task_process} (a) presents an overview of the MAV as it carries out the scanning operation. 
Fig. \ref{fig: task_process} (b1) illustrates the real-world scenario corresponding to this task. 
Fig. \ref{fig: task_process} (b2) depicts the task map generated during the exploration phase, where black voxels denote obstacles and green voxels represent real-time sensor perceptions. 
Additionally, Fig. \ref{fig: task_process} (b3) shows the orientation of the MAV during the scanning procedure. 
Various perspectives of the entire process are provided in Figs. \ref{fig: task_process} (c1) and (c2). 
These figures demonstrate both the reference path (green) and the actual scanning path (red), which was optimized through the motion planning algorithm.

When the MAV received the scanning task, it first carried out exploration, modelling the obstacles in the environment where the instance is located using the principle of ray-casting during flight. After the exploration, the reference scan path was checked for collisions with the obstacle map, and the A* algorithm was used to find feasible paths for the parts of the path blocked by obstacles. Finally, the scanning was performed with the orientation of the MAVs pointing towards the instance. After completing all procedures, the MAV returned to its inspection mode and proceeded to the next inspection target point.

\subsubsection{Evaluation on Visual Data Qualities} \label{section: dq}
The quality of images collected by MAVs is crucial for downstream processes like reconstructing structural models. This section evaluates the image quality captured during SHM tasks using the Natural Image Quality Evaluator (NIQE) \cite{mittal2012making}, which assesses image quality without reference images.
First, the local contrast of the image is normalized to calculate the Mean Subtracted Contrast Normalized (MSCN) coefficients:
\begin{equation}
    \mathrm{MSCN}(i,j)=\frac{I(i,j)-\mu(i,j)}{\sigma(i,j)+C}
\end{equation}

where $I(i,j)$ represents the value at pixel $(i,j)$, $\mu(i,j)$ and $\sigma(i,j)$ are the local mean and standard deviation, respectively, and $C$ is a constant.
Next, the NIQE score is determined by calculating the Mahalanobis distance between the test image's feature vector and the natural scene model:
\begin{equation}
    NIQE=\sqrt{(\mathbf{x}-\mu)^T\boldsymbol{\Sigma}^{-1}(\mathbf{x}-\mu)}
\end{equation}
where \(\mathbf{x}\) is the vector of the test image, and \(\mu\) and \(\mathbf{\Sigma}\) are the mean vector and covariance matrix of the natural scene model.
To make the NIQE score more intuitive, it is normalised from 0 to 1, with 0 representing the highest quality and 1 the lowest:
\begin{equation}
\mathrm{NIQE_{norm}} = \frac{\mathrm{NIQE} - \mathrm{NIQE_{\text{min}}}}{\mathrm{NIQE_{\text{max}}} - \mathrm{NIQE_{\text{min}}}}
\end{equation}

\begin{figure}
    \centering
    \includegraphics[width=1\linewidth]{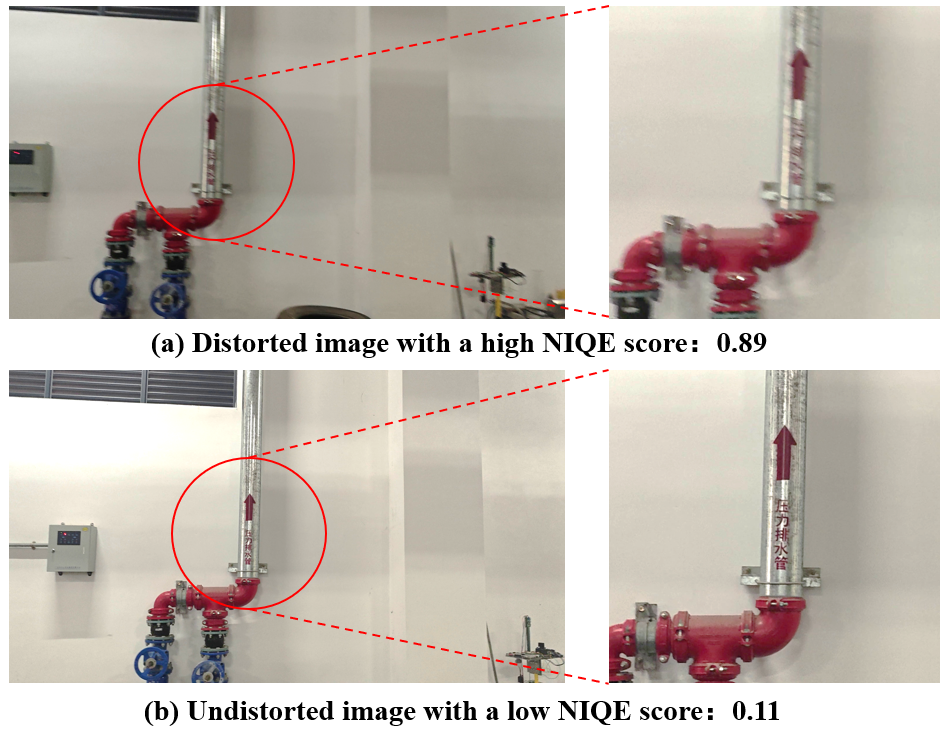}
    \caption{NIQE scores of the captured image}
    \label{fig:NIQE}
\end{figure}
The comparison with high-scored and low-scored data is shown in Fig.\ref{fig:NIQE}. 
The image with distortions (Fig.\ref{fig:NIQE}-a) receives a NIQE score close to 1, while a normal image (Fig.\ref{fig:NIQE}-b) has a much lower NIQE score. 
To enhance data usability and improve the quality of 3D reconstruction, we set a threshold for the NIQE score, $S_{dis}$, after data collection. 
Any data over $S_{dis}$ is considered distorted and is automatically removed from the dataset.

\subsubsection{Evaluation on Structure Reconstruction from 3D GS} \label{section: 3drs}
\begin{figure*}
    \centering
    \includegraphics[width=1\linewidth]{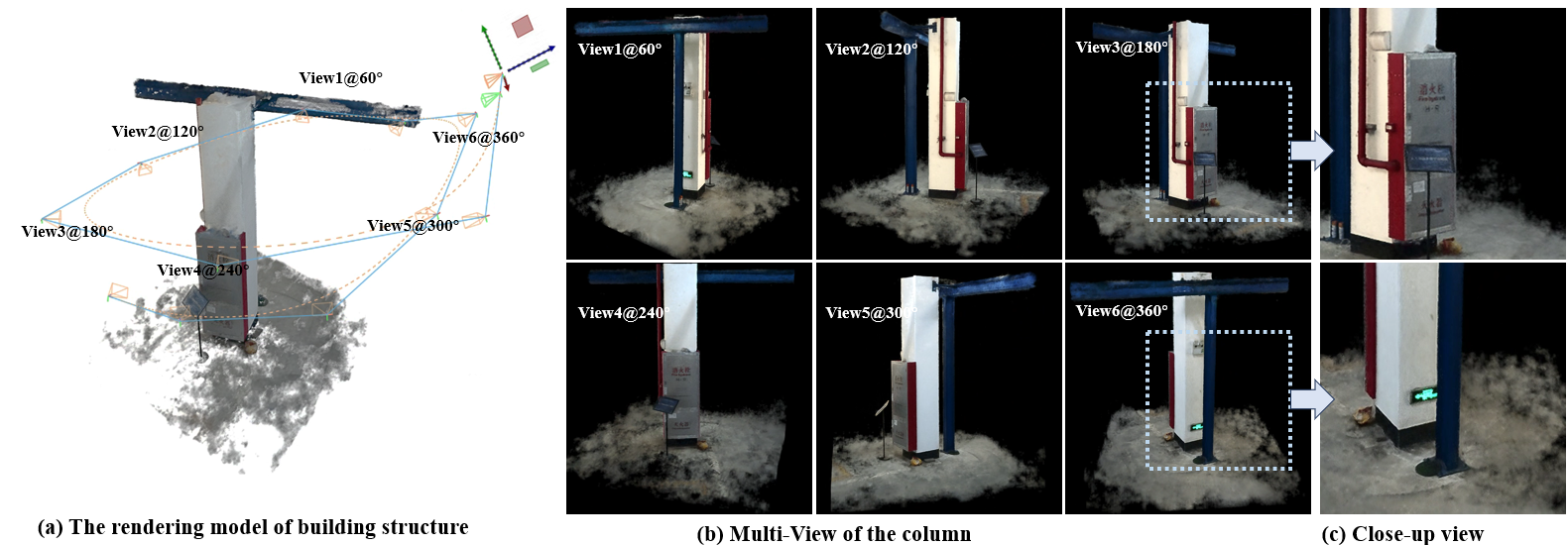}
    \caption{Multi-angle visual inspection on column via 3D GS }
    \label{fig:multi-angle}
\end{figure*}

Following the evaluation of image quality, all high-quality image data were prepared for structural reconstruction. 
We employed a novel image-based 3D modelling technique, 3D Gaussian Splatting (GS) \cite{kerbl20233d}, to generate real-time rendering models for each structure. 
In comparison to direct dense point cloud modelling, this method facilitates the real-time rendering of high-fidelity images from any viewpoint and distance, thereby enabling detailed and efficient visual inspection of the structures. 
Fig.\ref{fig:multi-angle} illustrates visualizations of inspections from various orientations.
To assess the quality of each rendered model, we utilized the Peak Signal-to-Noise Ratio (PSNR) metric, which is commonly used in computer graphics to measure image fidelity. 
The PSNR indicates the similarity between the rendered images and the original images; a higher PSNR value denotes greater fidelity \cite{hu2024high}. 
For the model shown in Fig.\ref{fig:multi-angle}, the PSNR is 32 dB, while the average PSNR for the other datasets in this study is 30.8 dB.

\subsection{Experiment 2: Evaluation of MAVs Manoeuvrability} \label{section: mp}
\subsubsection{Evaluation of Planning Performance}
In this experiment, we evaluated the MAV planner's performance across several key metrics under different conditions and summarized the factors affecting the planner. 
The test results are reported in Table \ref{tab:planning_performance}. 
The metrics include:
\(\mathbf{T_G}\) (ms): Computation time for initial path generation.
\(\mathbf{T_{Opt}}\) (ms): Optimization time.
\(\mathbf{T_{A*}}\) (ms): Search time for obstacle avoidance gradients.
\(\mathbf{D}\) (m): Euclidean distance between the target point and the MAV.
\(\mathbf{L_{final}}\) (m): Actual path length planned by the planner.

\begin{table}[h]
    \caption{Comparison of Path Planning Performance}
    \label{tab:planning_performance}
    \centering
    \begin{tabular}{cccccc}
            \toprule
            {\textbf{Case}}   & $\mathbf{T_G}$  & $\mathbf{T_{Opt}}$  & $\mathbf{T_{A*}}$ &{$\mathbf{D}$} &{$\mathbf{L_{final}}$} \\ 
            \midrule
            No Obstacle      & 0.069    & 0.181       & 0          &2.0   &1.998\\
            No Obstacle      & 0.109    & 0.353       & 0          &4.0   &3.989\\
            No Obstacle      & 0.113    & 0.872       & 0          &6.0   &6.002\\
            Small Obstacle   & 0.112    & 1.093       & 2.7        &6.0   &7.202\\
            Large Obstacle   & 0.113    & 14.268      & 19.24      &6.0   &7.969\\
            \bottomrule
        \end{tabular}
\end{table}

We first evaluated the planner's performance for targets at varying distances without considering obstacles. Initial paths were generated using the Minisnap algorithm. 
As shown in the table \ref{tab:planning_performance}, both generation and optimization times increased with distance due to the higher number of path points in longer paths, resulting in greater computational complexity.
Next, we assessed the MAVs' performance in obstacle avoidance with obstacles of varying sizes by adjusting the dilation coefficients to simulate dense environments. 
Results indicate that larger obstacles lead to longer search times for obstacle avoidance gradients. This occurs because the A* algorithm, used by the planner, requires searching more grid cells to navigate around larger obstacles, thereby increasing optimization time.
The planner's efficiency decreases with increasing target distance and obstacle size, underscoring the need to consider these factors in planning scenarios. 
Overall, our planning method maintains a fast computational speed of 14ms and achieves a 100\% success rate, even with long distances and obstacles.

\begin{figure}[h]
    \centering
    \includegraphics[width=1\linewidth]{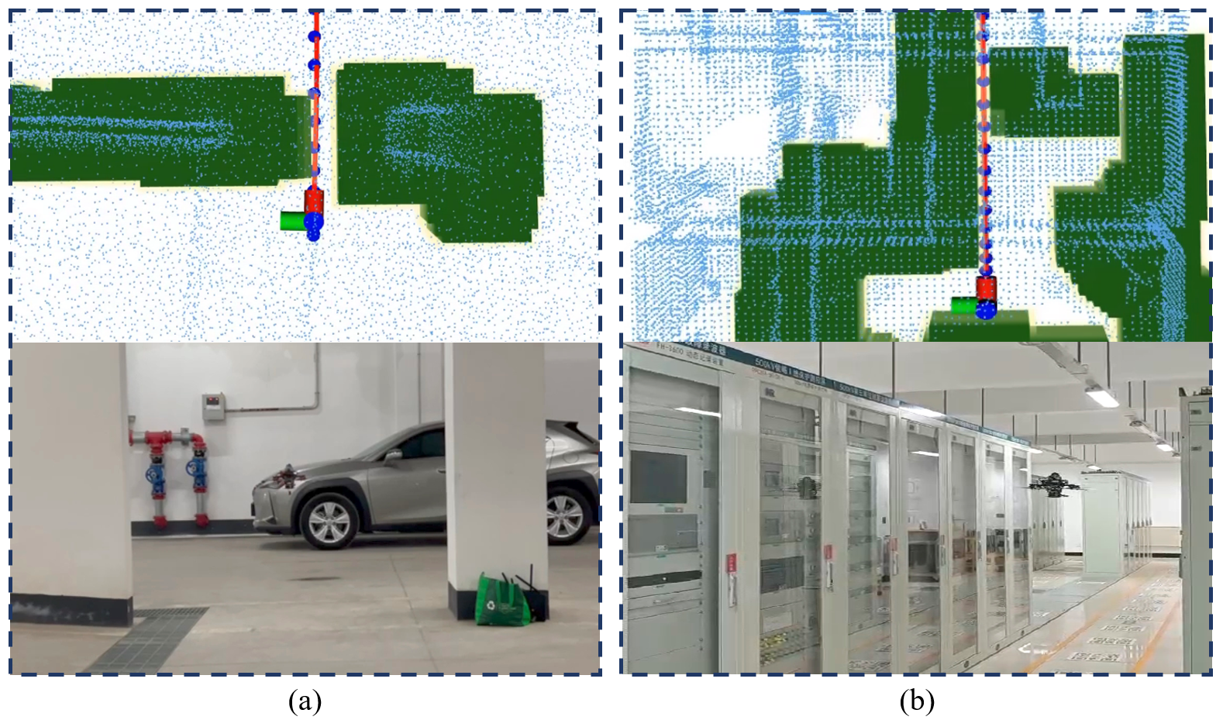}
    \caption{Demonstration of Planning Performance in Narrow Areas (a) Top: Planning Results with a Two-Meter Interval. Bottom: Actual Flight Performance. (b) Top: Planning Results with a 1.4-Meter Interval. Bottom: Actual Flight Performance.}
    \label{fig: narrow_area}
\end{figure}

\subsubsection{MAVs' Manoeuvrability in Narrow Space}
This section evaluates the planner's performance in two narrow scenarios, illustrated in Fig. \ref{fig: narrow_area}. 
To simulate the MAV navigating through tight spaces, we set a large inflation coefficient. 
Fig. \ref{fig: narrow_area}(a) shows a real-world obstacle spacing of 2m, where a 0.7m inflation coefficient reduces the passable space to 0.6m. 
Similarly, Fig. \ref{fig: narrow_area}(b) depicts a shallow corridor of 1.4m, leaving the MAV with only 0.3m to pass through. Despite these constraints, the planning results indicate that the planner performs effectively in such narrow environments.

\subsubsection{Evaluation of MAVs Motion Accuracy} \label{section: mt}
\begin{figure}[h]
    \centering
    \includegraphics[width=0.5\textwidth]{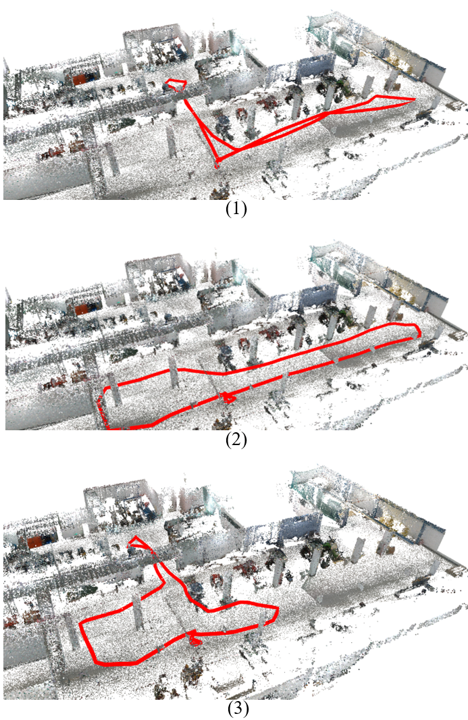}
    \caption{Demonstration on experiment paths (a) path 1 (b) path 2 (c) path 3}
    \label{fig: motion_exp_path}
\end{figure}

This experiment evaluates the path-tracking error between the planned trajectory and the actual flying trajectory during flying. 
Three different paths were selected for testing, each with planned speeds of 1.0, 1.75, and 2.5 m/s for flight. 
Odometry and controller outputs were recorded to calculate Absolute Position Error (APE) and Relative Position Error (RPE), along with corresponding Translation Error (Trans) and Root Mean Square Error (RMSE) values. 
Experimental results are listed in Table \ref{tab:motion_exp_path}, and flight path is shown in Fig. \ref{fig: motion_exp_path}.

\begin{table}[h]
    \small
    \caption{Comparison of Motion Accuracy at Different Speeds}
    \label{tab:motion_exp_path}
    \centering
    \begin{tabular}{l|c|cc|ccp{7.5cm}}
            \toprule
            \multirow{2}{*}{\textbf{Path}}   & \multirow{2}{*}{\textbf{vel.(m/s)}}  &\multicolumn{2}{c|}{{\textbf{APE}}} & \multicolumn{2}{c}{\textbf{RPE}} \\ 
                                            
                                             &    &$Max$  & $RMSE$ &$Max$ & $RMSE$ \\ 
            
            \midrule
            \multirow{3}{*}{\textit{path 1}}   & 1.0  &0.284  & 0.122 &0.515  & 0.095   \\
                                      &1.75  &0.470  & 0.114 &0.500   & 0.106   \\
                                      & 2.5  &0.345  & 0.130 &0.498  & 0.135   \\
            \hline

            \multirow{3}{*}{\textit{path 2}}   & 1.0  &0.351  & 0.162 &0.290 & 0.065  \\
                                      &1.75  &0.244  & 0.130 &0.356 & 0.083  \\
                                      & 2.5  &0.206  & 0.126 &0.507 & 0.186  \\
            \hline

            \multirow{3}{*}{\textit{path 3}}   & 1.0  &0.331  & 0.163 &0.458 & 0.091  \\
                                      &1.75  &0.327  & 0.149 &0.404 & 0.126  \\
                                      & 2.5  &0.361  & 0.203 &0.478 & 0.179  \\
            \bottomrule
        \end{tabular}
\end{table}

The results show that the UAV's path-tracking accuracy varies with different flying speeds. At lower speeds (1.0 m/s), the performance of both APE RMSE and RPE RMSE tends to be more stable and accurate across different paths. However, as the speed increases to 2.5 m/s, there is a noticeable deterioration in accuracy. Path 3, in particular, exhibits the highest errors at the highest speed, with an APE RMSE of 0.203 and an RPE RMSE of 0.179. This suggests that higher speeds present a challenge for the motion-tracking system, leading to increased errors.
Path 2 shows an interesting trend where the APE RMSE improves slightly at medium speed (1.75 m/s), but the RPE RMSE increases significantly at the highest speed. Specifically, the APE RMSE decreases from 0.162 at 1.0 m/s to 0.130 at 1.75 m/s, indicating better path-tracking at medium speed. However, the RMSE of RPE jumps from 0.065 at 1.0 m/s to 0.186 at 2.5 m/s, indicating higher relative position errors at the highest speed.
In scanning tasks, MAVs will operate at a velocity of 0.5 m/s, which ensures highly precise motion tracking and stable image data acquisition.

\begin{figure}[h]
    \centering
    \includegraphics[width=1\linewidth]{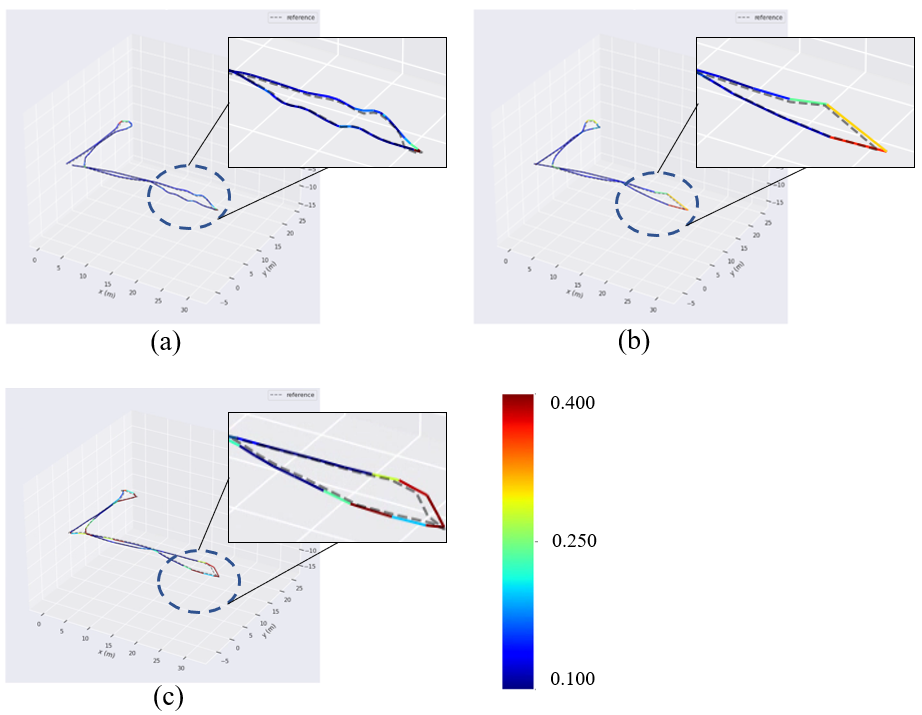}
    \caption{Distribution of RPE Along the Same Path at Different Planning Velocities (a) 1.0 m/s (b) 1.75 m/s (c) 2.5 m/s}
    \label{fig: evo_error_rpe}
\end{figure}
\subsubsection{Analysis of Motion Error}

We conducted a quantitative evaluation of the error sources affecting MAVs' motion accuracy by calculating the Relative Pose Error (RPE) for executing planned paths at varying speeds (1.0, 1.75, and 2.5 m/s). Using the EVO tool, we visualized the discrepancies between the planned and actual trajectories, as shown in Fig. \ref{fig: evo_error_rpe}. 
The results indicate that higher motion errors predominantly occurred at the turning segments of the paths, likely due to the increased complexity of maintaining accurate control during rapid directional changes, especially at higher speeds. 
Additionally, experiments conducted at higher speeds (1.75 and 2.5 m/s) revealed more pronounced errors during the straight segments, with the highest error observed at 2.5 m/s. 
This suggests that as speed increases, the ability to maintain precise trajectory tracking is reduced.

\subsection{Discussion} \label{section: Discussion}
\textbf{Fast and high-fidelity inspection}: 
Combining the emerging neural reconstruction approach with an autonomous MAV system significantly enhances the speed and fidelity of the inspection operation. 
As illustrated in Fig. \ref{fig:multi-angle}, our method achieves high-fidelity reconstruction of structure geometry and appearance, capturing fine details from all view angles. 
This ability critically relied on the high motion accuracy and flying stability of MAVs, which enabled them to acquire high-resolution and quality images for reconstruction.
This improvement is further supported by the MAV’s instance-aware perception and obstacle-free motion planning capabilities, enabling fast and accurate instance-level inspection of key structures, even in complex and denied construction environments (illustrated in Figs. \ref{fig:total_flight} and \ref{fig: task_process}).
The integration of these advanced methods allows for efficient and high-precision modelling, ensuring comprehensive environmental information and seamless transitions between flying and inspection tasks, as evidenced by the successful implementation of the system in the demonstration video.

\begin{figure}[ht]
    \centering
    \includegraphics[width=1\linewidth]{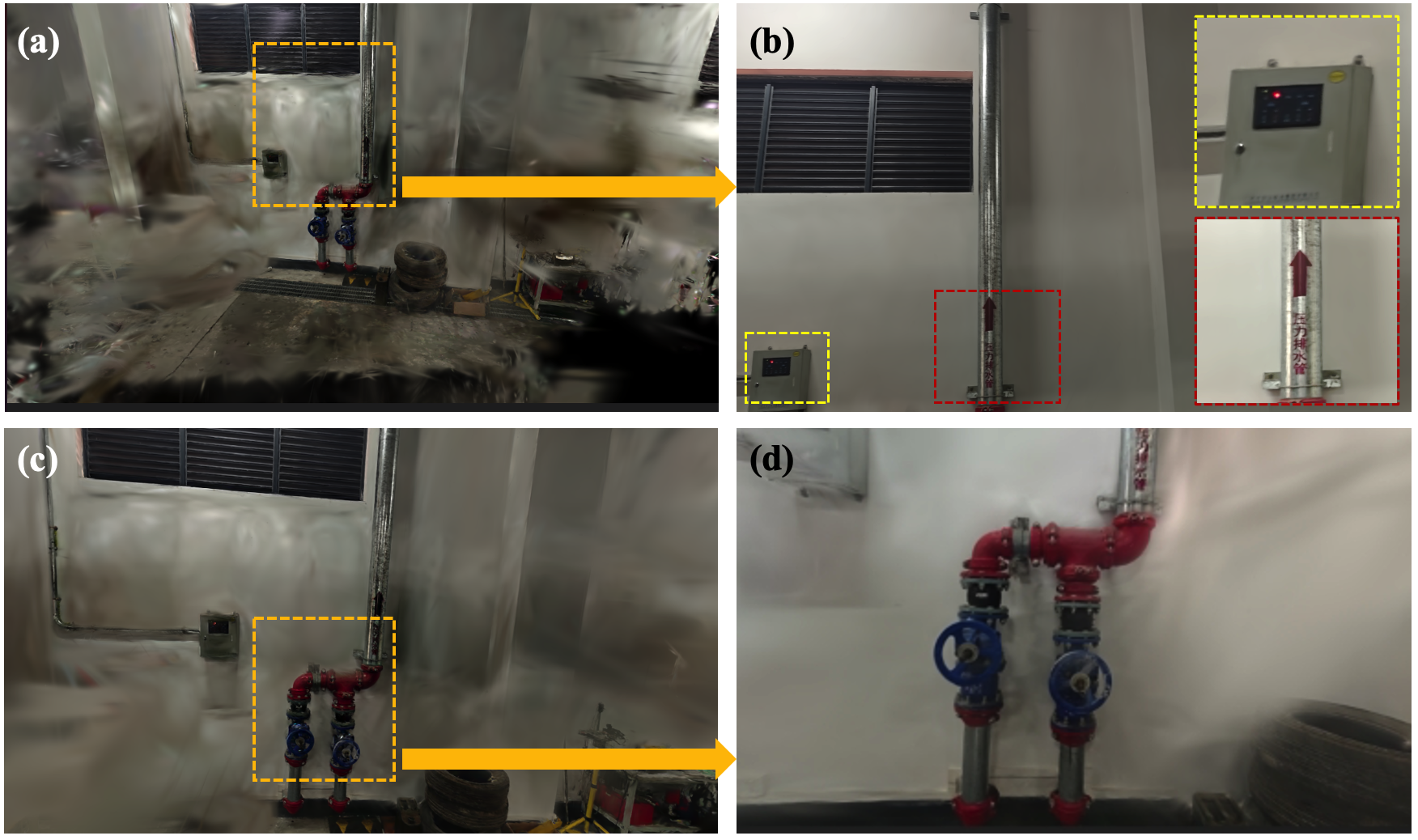}
    \caption{Reconstructed . (a) and (c) show 3D visualisation of reconstructed structures; (b) and (d) are detailed rendering images from certain view angle}
    \label{fig: discuss 1}
\end{figure}

we should also note that the quality of the reconstruction results is also influenced by the properties of the scenes. As illustrated in Fig. \ref{fig:multi-angle}, structures with rich features across different scanning trajectories maintain high-quality reconstructions. However, when dealing with scenes that consist largely of non-textured structures, such as walls, the reconstruction quality tends to be lower, as shown in Fig. \ref{fig: discuss 1}. In contrast, regions with relatively rich textures, as depicted in Figs. \ref{fig: discuss 1} (b) and (c), consistently yield higher-quality model reconstructions.

\textbf{Improved efficiency and safety}:
Remote control of drones in denied and narrow indoor environments presents significant challenges, even for experienced operators. 
Our method addresses these challenges by equipping MAVs with precise localization, obstacle perception, and motion planning capabilities specifically designed for such constrained spaces. 
This enables the MAV to operate safely and efficiently, as illustrated in Figs. \ref{fig: task_process}, \ref{fig: narrow_area}, and \ref{fig: motion_exp_path}. 
By maintaining a between-instance flying speed of 2m/s and an instance scanning speed of 0.5m/s, our system strikes an effective balance between accuracy and efficiency, resulting in high-quality neural renderings in the final output.
This novel design requires less human intervention during operation, leading to a safer yet more efficient inspection paradigm for construction. 
However, it is important to note that our method has a high computational load throughout the entire operation, and its performance is further constrained by the limited flying time of MAVs, which currently allows for only 10-12 minutes of operation, covering 3-4 instances, both aspects can be optimized in future work.

\section{Conclusion} \label{section: conclusion}
This study presents a novel framework for MAV inspections in complex indoor environments, featuring a hierarchical perception and planning system for optimised task paths and advanced localisation and motion planning. 
Integrated with neural reconstruction technology, the framework enables comprehensive 3D reconstruction of building structures. 
Empirical validation in a 4,000 $m^2$ underground facility demonstrated exceptional performance, with a 100\% success rate in both autonomous inspection and motion planning, and high fidelity rendering models using the 3DGS model. 
Overall, this study makes several key contributions to the field of autonomous inspection in complex indoor environments. 
The proposed method enhances automatic inspection capabilities through an instance-aware planning strategy, while the designed perception and motion generation framework ensures high self-localisation accuracy without relying on GPS and provides robust motion control in challenging conditions. 
In addition, the integration of a 3D Gaussian Splatting (3D GS) reconstruction approach allows effective management of inspection data and enables detailed visual observations from different viewpoints. 
However, the method has certain limitations, including a high computational load, which may limit its application to larger buildings or long-distance inspections. 
The path-tracking experiment found an inverse relationship between speed and accuracy, suggesting that higher speeds may lead to increased tracking errors, which could affect the versatility of the system. 
In addition, the neural reconstruction method faces challenges in aligning image data with minimal visual features, such as clean, white walls.



\bibliographystyle{IEEEtran}
\bibliography{root}

\begin{thebibliography}{10}
\providecommand{\url}[1]{#1}
\csname url@samestyle\endcsname
\providecommand{\newblock}{\relax}
\providecommand{\bibinfo}[2]{#2}
\providecommand{\BIBentrySTDinterwordspacing}{\spaceskip=0pt\relax}
\providecommand{\BIBentryALTinterwordstretchfactor}{4}
\providecommand{\BIBentryALTinterwordspacing}{\spaceskip=\fontdimen2\font plus
\BIBentryALTinterwordstretchfactor\fontdimen3\font minus \fontdimen4\font\relax}
\providecommand{\BIBforeignlanguage}[2]{{%
\expandafter\ifx\csname l@#1\endcsname\relax
\typeout{** WARNING: IEEEtran.bst: No hyphenation pattern has been}%
\typeout{** loaded for the language `#1'. Using the pattern for}%
\typeout{** the default language instead.}%
\else
\language=\csname l@#1\endcsname
\fi
#2}}
\providecommand{\BIBdecl}{\relax}
\BIBdecl

\bibitem{pena2024uav}
F.~L. Pe{\~n}a, {\'A}.~D. D{\'\i}az, F.~Orjales, and J.~L. Pita, ``An uav system for visual inspection and wall thickness measurements in ship surveys,'' \emph{Measurement}, p. 115262, 2024.

\bibitem{wang2024rapid}
F.~Wang, Y.~Zou, X.~Chen, C.~Zhang, L.~Hou, E.~del Rey~Castillo, and J.~B. Lim, ``Rapid in-flight image quality check for uav-enabled bridge inspection,'' \emph{ISPRS Journal of Photogrammetry and Remote Sensing}, vol. 212, pp. 230--250, 2024.

\bibitem{zhang2024developing}
N.~Zhang, Y.~Pan, Y.~Jin, P.~Jin, K.~Hu, X.~Huang, and H.~Kang, ``Developing a flying explorer for autonomous digital modelling in wild unknowns,'' \emph{Sensors}, vol.~24, no.~3, p. 1021, 2024.

\bibitem{pan2024pheno}
Y.~Pan, K.~Hu, T.~Liu, C.~Chen, and H.~Kang, ``Pheno-robot: An auto-digital modelling system for in-situ phenotyping in the field,'' \emph{arXiv preprint arXiv:2402.09685}, 2024.

\bibitem{yu2022uav}
C.~Yu, Y.~Yang, Y.~Cheng, Z.~Wang, M.~Shi, and Z.~Yao, ``Uav-based pipeline inspection system with swin transformer for the east,'' \emph{Fusion engineering and design}, vol. 184, p. 113277, 2022.

\bibitem{zhang2024reactive}
R.~Zhang, G.~Hao, K.~Zhang, and Z.~Li, ``Reactive uav-based automatic tunnel surface defect inspection with a field test,'' \emph{Automation in Construction}, vol. 163, p. 105424, 2024.

\bibitem{stefan2023multi}
I.~Stefan, C.~Bojan, G.~Luka, and M.~Lea, ``Multi-uav trajectory planning for 3d visual inspection of complex structures [j],'' \emph{Automation in Construction}, vol. 147, p. 104709, 2023.

\bibitem{gao2023uav}
C.~Gao, X.~Wang, R.~Wang, Z.~Zhao, Y.~Zhai, X.~Chen, and B.~M. Chen, ``A uav-based explore-then-exploit system for autonomous indoor facility inspection and scene reconstruction,'' \emph{Automation in Construction}, vol. 148, p. 104753, 2023.

\bibitem{tan2021automatic}
Y.~Tan, S.~Li, H.~Liu, P.~Chen, and Z.~Zhou, ``Automatic inspection data collection of building surface based on bim and uav,'' \emph{Automation in Construction}, vol. 131, p. 103881, 2021.

\bibitem{dianovsky2023electromagnetic}
R.~Dianovsk{\`y}, P.~Pecho, P.~Vel'k{\`y}, and M.~Hr{\'u}z, ``Electromagnetic radiation from high-voltage transmission lines: Impact on uav flight safety and performance,'' \emph{Transportation research procedia}, vol.~75, pp. 209--218, 2023.

\bibitem{pan2024novel}
Y.~Pan, K.~Hu, H.~Cao, H.~Kang, and X.~Wang, ``A novel perception and semantic mapping method for robot autonomy in orchards,'' \emph{Computers and Electronics in Agriculture}, vol. 219, p. 108769, 2024.

\bibitem{aliyari2022hazards}
M.~Aliyari, B.~Ashrafi, and Y.~Z. Ayele, ``Hazards identification and risk assessment for uav--assisted bridge inspections,'' \emph{Structure and Infrastructure Engineering}, vol.~18, no.~3, pp. 412--428, 2022.

\bibitem{li2023automatic}
R.~Li, J.~Yu, F.~Li, R.~Yang, Y.~Wang, and Z.~Peng, ``Automatic bridge crack detection using unmanned aerial vehicle and faster r-cnn,'' \emph{Construction and Building Materials}, vol. 362, p. 129659, 2023.

\bibitem{mu2023adaptive}
Z.~Mu, Y.~Qin, C.~Yu, Y.~Wu, Z.~Wang, H.~Yang, and Y.~Huang, ``Adaptive cropping shallow attention network for defect detection of bridge girder steel using unmanned aerial vehicle images,'' \emph{Journal of Zhejiang University-SCIENCE A}, vol.~24, no.~3, pp. 243--256, 2023.

\bibitem{cui2023skip}
J.~Cui, Y.~Qin, Y.~Wu, C.~Shao, and H.~Yang, ``Skip connection yolo architecture for noise barrier defect detection using uav-based images in high-speed railway,'' \emph{IEEE Transactions on Intelligent Transportation Systems}, 2023.

\bibitem{yu2023amcd}
L.~Yu, E.~Yang, C.~Luo, and P.~Ren, ``Amcd: an accurate deep learning-based metallic corrosion detector for mav-based real-time visual inspection,'' \emph{Journal of Ambient Intelligence and Humanized Computing}, vol.~14, no.~7, pp. 8087--8098, 2023.

\bibitem{demkiv2021application}
L.~Demkiv, M.~Ruffo, G.~Silano, J.~Bednar, and M.~Saska, ``An application of stereo thermal vision for preliminary inspection of electrical power lines by mavs,'' in \emph{2021 Aerial Robotic Systems Physically Interacting with the Environment (AIRPHARO)}.\hskip 1em plus 0.5em minus 0.4em\relax IEEE, 2021, pp. 1--8.

\bibitem{ortiz2016vision}
A.~Ortiz, F.~Bonnin-Pascual, E.~Garcia-Fidalgo, and J.~P. Company-Corcoles, ``Vision-based corrosion detection assisted by a micro-aerial vehicle in a vessel inspection application,'' \emph{Sensors}, vol.~16, no.~12, p. 2118, 2016.

\bibitem{bonnin2019reconfigurable}
F.~Bonnin-Pascual, A.~Ortiz, E.~Garcia-Fidalgo, and J.~P. Company-Corcoles, ``A reconfigurable framework to turn a mav into an effective tool for vessel inspection,'' \emph{Robotics and Computer-Integrated Manufacturing}, vol.~56, pp. 191--211, 2019.

\bibitem{dong2024neural}
Z.~Dong, W.~Lu, and J.~Chen, ``Neural rendering-based semantic point cloud retrieval for indoor construction progress monitoring,'' \emph{Automation in Construction}, vol. 164, p. 105448, 2024.

\bibitem{cui20243d}
D.~Cui, W.~Wang, W.~Hu, J.~Peng, Y.~Zhao, Y.~Zhang, and J.~Wang, ``3d reconstruction of building structures incorporating neural radiation fields and geometric constraints,'' \emph{Automation in Construction}, vol. 165, p. 105517, 2024.

\bibitem{bachrach2011range}
A.~Bachrach, S.~Prentice, R.~He, and N.~Roy, ``Range--robust autonomous navigation in gps-denied environments,'' \emph{Journal of Field Robotics}, vol.~28, no.~5, pp. 644--666, 2011.

\bibitem{mohta2018fast}
K.~Mohta, M.~Watterson, Y.~Mulgaonkar, S.~Liu, C.~Qu, A.~Makineni, K.~Saulnier, K.~Sun, A.~Zhu, J.~Delmerico \emph{et~al.}, ``Fast, autonomous flight in gps-denied and cluttered environments,'' \emph{Journal of Field Robotics}, vol.~35, no.~1, pp. 101--120, 2018.

\bibitem{bi2017robust}
Y.~Bi, M.~Lan, J.~Li, K.~Zhang, H.~Qin, S.~Lai, and B.~M. Chen, ``Robust autonomous flight and mission management for mavs in gps-denied environments,'' in \emph{2017 11th Asian Control Conference (ASCC)}.\hskip 1em plus 0.5em minus 0.4em\relax IEEE, 2017, pp. 67--72.

\bibitem{younes2021optimal}
Y.~A. Younes and M.~Barczyk, ``Optimal motion planning in gps-denied environments using nonlinear model predictive horizon,'' \emph{Sensors}, vol.~21, no.~16, p. 5547, 2021.

\bibitem{shithil2022robust}
S.~M. Shithil, A.~A.~M. Faudzi, A.~Abdullah, N.~Islam, and S.~M. Saad, ``Robust sensor fusion for autonomous uav navigation in gps denied forest environment,'' in \emph{2022 IEEE 5th International Symposium in Robotics and Manufacturing Automation (ROMA)}.\hskip 1em plus 0.5em minus 0.4em\relax IEEE, 2022, pp. 1--6.

\bibitem{zhang2016easy}
W.~Zhang, J.~Qi, P.~Wan, H.~Wang, D.~Xie, X.~Wang, and G.~Yan, ``An easy-to-use airborne lidar data filtering method based on cloth simulation,'' \emph{Remote sensing}, vol.~8, no.~6, p. 501, 2016.

\bibitem{sarle1991finding}
W.~S. Sarle, ``Finding groups in data: An introduction to cluster analysis.'' 1991.

\bibitem{zhang1999flexible}
Z.~Zhang, ``Flexible camera calibration by viewing a plane from unknown orientations,'' in \emph{Proceedings of the seventh ieee international conference on computer vision}, vol.~1.\hskip 1em plus 0.5em minus 0.4em\relax Ieee, 1999, pp. 666--673.

\bibitem{xu2022fast}
W.~Xu, Y.~Cai, D.~He, J.~Lin, and F.~Zhang, ``Fast-lio2: Fast direct lidar-inertial odometry,'' \emph{IEEE Transactions on Robotics}, vol.~38, no.~4, pp. 2053--2073, 2022.

\bibitem{koide2021voxelized}
K.~Koide, M.~Yokozuka, S.~Oishi, and A.~Banno, ``Voxelized gicp for fast and accurate 3d point cloud registration,'' in \emph{2021 IEEE International Conference on Robotics and Automation (ICRA)}.\hskip 1em plus 0.5em minus 0.4em\relax IEEE, 2021, pp. 11\,054--11\,059.

\bibitem{zhou2020ego}
X.~Zhou, Z.~Wang, H.~Ye, C.~Xu, and F.~Gao, ``Ego-planner: An esdf-free gradient-based local planner for quadrotors,'' \emph{IEEE Robotics and Automation Letters}, vol.~6, no.~2, pp. 478--485, 2020.

\bibitem{shan2020lio}
T.~Shan, B.~Englot, D.~Meyers, W.~Wang, C.~Ratti, and D.~Rus, ``Lio-sam: Tightly-coupled lidar inertial odometry via smoothing and mapping,'' in \emph{2020 IEEE/RSJ international conference on intelligent robots and systems (IROS)}.\hskip 1em plus 0.5em minus 0.4em\relax IEEE, 2020, pp. 5135--5142.

\bibitem{saito2015precision}
T.~Saito and M.~Rehmsmeier, ``The precision-recall plot is more informative than the roc plot when evaluating binary classifiers on imbalanced datasets,'' \emph{PloS one}, vol.~10, no.~3, p. e0118432, 2015.

\bibitem{mittal2012making}
A.~Mittal, R.~Soundararajan, and A.~C. Bovik, ``Making a “completely blind” image quality analyzer,'' \emph{IEEE Signal processing letters}, vol.~20, no.~3, pp. 209--212, 2012.

\bibitem{kerbl20233d}
B.~Kerbl, G.~Kopanas, T.~Leimk{\"u}hler, and G.~Drettakis, ``3d gaussian splatting for real-time radiance field rendering.'' \emph{ACM Trans. Graph.}, vol.~42, no.~4, pp. 139--1, 2023.

\bibitem{hu2024high}
K.~Hu, W.~Ying, Y.~Pan, H.~Kang, and C.~Chen, ``High-fidelity 3d reconstruction of plants using neural radiance fields,'' \emph{Computers and Electronics in Agriculture}, vol. 220, p. 108848, 2024.

\end{thebibliography}
\vfill

\end{document}